%% file: main.tex
\pdfoutput=1

\documentclass[11pt]{article}

\usepackage[final]{acl}

\usepackage{times}
\usepackage{latexsym}

\usepackage[T1]{fontenc}

\usepackage[utf8]{inputenc}

\usepackage{microtype}

\usepackage{inconsolata}

\usepackage{graphicx}

\usepackage{microtype}
\usepackage{graphicx}
\usepackage{subfigure}
\usepackage{booktabs}

\usepackage{amsmath}
\usepackage{amssymb}
\usepackage{mathtools}
\usepackage{amsthm}

\usepackage[capitalize,noabbrev]{cleveref}

\usepackage{xspace}

\usepackage{tikzducks}
\usepackage{lipsum}

\usepackage{graphicx}

\usepackage{enumitem}
\usepackage[export]{adjustbox}

\usepackage{mathtools}

\usepackage{multirow}

\usepackage{colortbl}

\usepackage{amssymb}
\usepackage{pifont}

\newcommand{\cmark}{\ding{51}}%
\newcommand{\xmark}{\ding{55}}%

\newcommand{\algname}{\textbf{C}ontext-\textbf{A}ware \textbf{M}olecular \textbf{T5}\xspace}

\newcommand{\Algname}{CAMT5\xspace}

\definecolor{LightGray}{gray}{0.95}

\newcommand{\stdv}[1]{\scriptsize~$\pm$~#1}

\definecolor{tablegreen}{rgb}{0.91, 0.94, 0.97}
\definecolor{SJViolet}{RGB}{105,100,171}
\definecolor{SJRed}{RGB}{237,109,107}

\definecolor{darkblue}{rgb}{0.098,0.3,0.9}
\definecolor{aliceblue}{rgb}{0.91, 0.94, 0.97}

\definecolor{cornellred}{rgb}{0.7, 0.11, 0.11}
\definecolor{cadmiumgreen}{rgb}{0.0, 0.42, 0.24}
\definecolor{Gray}{gray}{0.9}
\hypersetup{citecolor=MidnightBlue, linkcolor=BrickRed}
\hypersetup{
  linkcolor = cornellred,
  citecolor  = cadmiumgreen,
  colorlinks = true,
}

\theoremstyle{plain}

\theoremstyle{definition}

\theoremstyle{remark}

\title{Training Text-to-Molecule Models with Context-Aware Tokenization}

\author{
 \textbf{Seojin Kim\textsuperscript{1}$^*$},
 \textbf{Hyeontae Song\textsuperscript{2}$^*$$^{\dagger}$},
 \textbf{Jaehyun Nam\textsuperscript{3}},
 \textbf{Jinwoo Shin\textsuperscript{3}}
\\
 \textsuperscript{1}Seoul National University,
 \textsuperscript{2}Moloco Inc.,\\
 \textsuperscript{3}Korea Advanced Institute of Science and Technology (KAIST)\\
 \texttt{osikjs@snu.ac.kr, hyeontae3109@gmail.com,} \\
 \texttt{\{jaehyun.nam, jinwoos\}@kaist.ac.kr}
}

\begin{document}
\maketitle
\begin{abstract}
Recently, text-to-molecule models have shown great potential across various chemical applications, e.g., drug-discovery.  These models adapt language models to molecular data by representing molecules as sequences of atoms. However, they rely on atom-level tokenizations, which primarily focus on modeling local connectivity, thereby limiting the ability of models to capture the global structural context within molecules.
To tackle this issue, we propose a novel text-to-molecule model, coined \emph{\algname} (\textbf{\Algname}). Inspired by the significance of the substructure-level contexts in understanding molecule structures, e.g., ring systems, we introduce substructure-level tokenization for text-to-molecule models. Building on our tokenization scheme, we develop an importance-based training strategy that prioritizes key substructures, enabling \Algname to better capture the molecular semantics. Extensive experiments verify the superiority of \Algname in various text-to-molecule generation tasks. 
Intriguingly, we find that \Algname outperforms the state-of-the-art methods using only 2\% of training tokens. In addition, we propose a simple yet effective ensemble strategy that aggregates the outputs of text-to-molecule models to further boost the generation performance. Code is available at \url{https://github.com/Songhyeontae/CAMT5.git}.
\end{abstract}

\input{1_introduction}
\input{2_related_work}

\input{3_method}

\input{4_experiments}

\input{5_conclusion}

\section*{Limitations}
In this work, we mainly focus on improving the token space of text-to-molecule models, which is a crucial yet under-explored problem in text-to-molecule models. An interesting future direction would be applying our tokenization to train advanced text-to-molecule models, e.g., leveraging pseudo-data \citep{chen2024artificially}, diffusion-based generation \citep{chang2024ldmol}, and multi-task language modeling \citep{christofidellis2023unifying}, which are originally based on the previous atom-wise tokenization schemes, e.g., SMILES \citep{weininger1988smiles}. We believe that those works will further benefit from our carefully designed context-aware tokenization.

\section*{Acknowledgments}
This work was supported by the National Supercomputing Center with supercomputing resources including technical support KSC-2024-CRE-0362, and partly supported by Institute for Information \& communications Technology Promotion(IITP) grant funded by the Korea government(MSIT) (No.RS-2019-II190075 Artificial Intelligence Graduate School Program (KAIST); RS-2022-II220959, Few-shot Learning of Causal Inference in Vision and Language for Decision Making).

\bibliography{custom}

\appendix
\onecolumn
\input{6_appendix}

\end{document}

%% file: 1_introduction.tex
\renewcommand\thefootnote{\fnsymbol{footnote}}

\footnotetext{\ignorespaces* These authors contributed equally.}
\footnotetext{\ignorespaces$^{\dagger}$ Work done at KAIST.
}
\renewcommand\thefootnote{\arabic{footnote}}

\section{Introduction}
\label{sec:introduction}

 Discovering molecules that match desired language descriptions has been a long-standing goal in chemistry since it is an essential ingredient for practical deployments like drug-discovery and material design \citep{su2022molecular,gong2024text,li2024empowering}. However, achieving such text-to-molecule generation poses a challenge because of the different structural modalities of language and molecules. To address this challenge, researchers have explored the fine-tuning of pre-trained language models with additional molecular data \citep{christofidellis2023unifying, chen2024artificially}, which is inspired by the recent success of language models in leveraging various domain knowledge, including chemical concepts \citep{taylor2022galactica,yu2024llasmol}. Specifically, they treat each molecule as a sequence of tokens using string-based molecular representations such as SMILES \citep{weininger1988smiles} and SELFIES \citep{krenn2020self}. Intriguingly, they show that these molecule-aware language models, i.e., text-to-molecule models, can be obtained by learning the text-conditional molecule distribution based on treating atoms as tokens of language models \citep{edwards2022translation,pei2023biot5}.

However, it is yet underexplored \emph{which} tokenization strategy for molecules is more effective for text-to-molecule models. Current state-of-the-art approaches \citep{edwards2022translation,pei2023biot5} adopt atom-level tokenization, where each atom is represented as a single token within the model's token space \citep{christofidellis2023unifying,liu2023molxpt,pei2023biot5}. 
Even though they show remarkable performance as pioneering efforts, such atom-level tokenizations limit the models' ability to capture crucial global contextual patterns within molecules, focusing only on local connectivities \citep{xia2022mole,liu2024rethinking,luong2024fragment}. 
This leads to the question of \emph{how to tokenize molecules in a context-preserving manner to train text-to-molecule models more effectively.}

\input{figures/concept}

\noindent\textbf{Contribution.} {In this paper, we introduce a novel text-to-molecule model, coined \emph{\algname} (\textbf{\Algname}), by proposing a context-enriched motif-level token space. Specifically, we draw inspiration from the following chemical prior---the structural context of molecules is more effectively captured through their substructure-level, i.e., motif-level, characteristics rather than the atom-level attributes \citep{jin2018junction,jin2020hierarchical,zhang2021motif,kim2023fragment}. Consequently, we hypothesize that text-to-molecule models can be further improved by emphasizing the information on key motifs during the training phase. To this end, we propose a new motif-level tokens for text-to-molecule models and develop a novel training strategy that effectively leverages the relative importance of the individual motif-level tokens (see Figure~\ref{fig:concept} for description).}

In particular, we carefully design the motif-level tokenization for \Algname to additionally alleviate two drawbacks in the tokenization strategies of previous text-to-molecule models.
First, \Algname~always generates a \emph{valid} molecule, while MolT5 \citep{edwards2022translation} often generates an \emph{invalid} token sequences that do not correspond to any molecule. Second, each of our motif-level tokens has a \emph{unique} interpretation, while some of the tokens in BioT5 \citep{pei2023biot5} have \emph{multiple} interpretations, e.g., both an atom and the number of atoms in a ring preceding the token \citep{krenn2020self}, introducing semantic-level ambiguities for the models.



{Consequently, to leverage our motif-level tokens effectively, we propose an importance-based training approach that prioritizes key motifs. Specifically, each token is assigned to an importance value derived from its constituent atoms, and the training loss is adjusted by weighting it according to this pre-defined importance. This loss design is made possible by our carefully designed motif-level tokens, each representing a unified chemical context \citep{kim2023fragment,luong2024fragment}, unlike atom-wise tokenizations \citep{weininger1988smiles,krenn2020self} in previous text-to-molecule models.}

We verify our method's effectiveness on popular benchmarks, e.g., ChEBI-20 \citep{edwards2021text2mol}. On ChEBI-20, the state-of-the-art results are achieved using only 2\% of the training tokens required by the previous best-performing baseline, BioT5 \citep{pei2023biot5}.\footnote{The performance of BioT5 \citep{pei2023biot5} benefits from additional non-public high-quality molecular pre-training data, which is not available for us.} Specifically, \Algname improves the ratio of molecules that exactly match the description (Exact; higher is better) by 0.413 $\rightarrow$ 0.430, and those similar to the description (RDK; higher is better) by 0.801 $\rightarrow$ 0.840. 
We also show that a simple ensemble strategy utilizing \Algname further improves the overall performance, e.g., Exact by 0.430 $\rightarrow$ 0.472. Finally, we verify \Algname's effectiveness in molecule modification.

%% file: figures/concept.tex
\begin{figure*}[t]
\centering
\includegraphics[width=\linewidth]{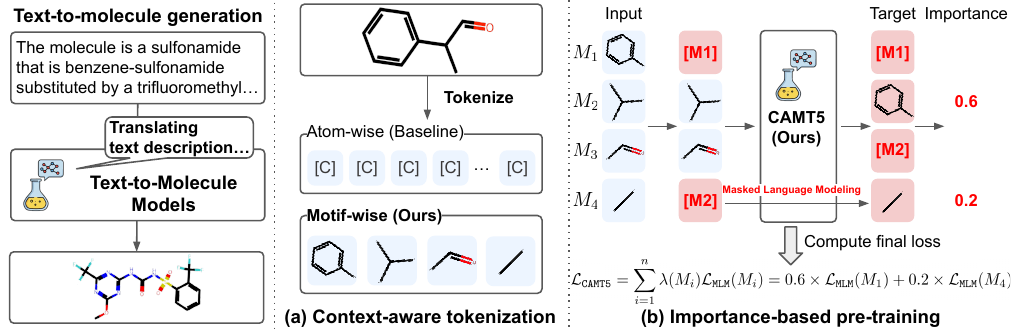}
\caption{
An overview of our proposed method. (a) Context-aware moecule tokenization: we train \Algname with a motif-level token sapce. (b) Importance-based pre-training: we priortize key motifs during pre-training.
}
\label{fig:concept}
\end{figure*}

%% file: 2_related_work.tex
\section{Related works}
\label{sec:related_work}

\textbf{Text-to-molecule models.} {Inspired by the recent advancements in language models \citep{raffel2020exploring,taylor2022galactica, openai2023gpt4}, significant efforts have been made to adapt these models for molecular applications, leading to the development of molecule-aware language models, i.e., text-to-molecule models \citep{edwards2021text2mol, edwards2022translation, christofidellis2023unifying, liu2023molxpt, pei2023biot5, chen2024artificially}. These approaches typically involve fine-tuning pre-trained language models, such as T5 \citep{raffel2020exploring}, using molecular data by representing molecules as sequences of atom-level tokens. For instance, existing models \citep{edwards2022translation,christofidellis2023unifying,pei2023biot5} leverage molecular representations like SMILES \citep{weininger1988smiles} or SELFIES \citep{krenn2020self}, which encode molecules as atom-level token sequences for text-to-molecule frameworks. As a result, they primarily focus on capturing local atom-wise connectivity, while overlooking the crucial global structural context of molecules \citep{zhang2021motif,xia2022mole,kim2023fragment,luong2024fragment}.}

{In addition, text-to-molecule models based on atom-level tokenizations come with additional drawbacks. First, SMILES-based models often generate \emph{invalid} token sequences that violate the grammar \citep{weininger1988smiles}, which do not correspond to valid molecules. Second, SELFIES-based models introduce semantic-level ambiguities, i.e., \emph{degeneracy}, in token interpretations \citep{krenn2020self}, leading to sub-optimal performance in modeling the token distribution. For example, the `$\mathtt{[O]}$' token can be interpreted completely differently: an oxygen atom or an indicator of a ring system comprising six atoms preceding this token. To address the aforementioned limitations, we carefully design our context-enriched motif-level tokens, ensuring \emph{validity} of the generated token sequences and \emph{non-degeneracy} in token interpretations (see Table~\ref{tab:advantages}).}

\input{tables/advantages}

\noindent\textbf{Context-aware molecule learning.} {Recent studies in the molecular domain have explored the concept of \emph{context-aware} learning for molecules. For example, \citet{zhang2021motif,kim2023fragment,luong2024fragment} propose self-supervised learning frameworks that leverage motif-level context to derive chemically meaningful molecular embeddings.
A notable approach in this line of work is context-aware molecule generation \citep{jin2018junction,jin2020hierarchical,kong2022molecule,geng2023de}, which focuses on learning the distribution of motifs instead of individual atoms with specialized architectures. Intriguingly, they show superior performance in molecule generation by incorporating contextual patterns of motifs within molecules. In particular, t-SMILES \citep{wu2024t} introduces a linearized representation of motifs using full binary tree structures. Therefore, they require additional grammar tokens to describe the full binary tree structures. In contrast, our motif tokens do not require any grammar tokens, enabling \Algname to concentrate solely on learning the relationships between motifs without being constrained by grammar representations (see Appendix~\ref{sup:t-smiles} for experimental comparison).}

%% file: tables/advantages.tex
\begin{table}[t]

\small

\begin{center}
\begin{tabular}{l|c|cc}
\toprule

Method & Token & Validity & Non-degeneracy\\\midrule
MolT5 & {Atom} &\textcolor{SJRed}{\xmark} & \textcolor{darkblue}\cmark\\
BioT5  & {Atom} &\textcolor{darkblue}\cmark & \textcolor{SJRed}{\xmark} \\\midrule
\rowcolor{tablegreen}\textbf{\Algname (Ours)} & {Motif} & \textcolor{darkblue}{\cmark} & \textcolor{darkblue}\cmark\\

\bottomrule

\end{tabular}
\end{center}
\vspace{-0.15in}
\caption{Comparison of our molecular tokens with previous text-to-molecule models. Token denotes the information encoded in a single token. We mark Validity if a sequence of tokens always represents a valid molecule, and we mark Non-degeneracy if a single token corresponds to a unique molecular interpretation.
}
\label{tab:advantages}
\vspace{-0.2in}
\end{table}

%% file: 3_method.tex
\section{Method}
\label{sec:method}

In Section~\ref{subsec:overview}, we explain an overview of our problem of interest. In Section~\ref{subsec:camt5}, we provide the description of our context-aware text-to-molecule model, \Algname. In Section~\ref{subsec:ensemble}, we describe our confidence-based ensemble strategy.

\subsection{Problem description} 
\label{subsec:overview}
We define our problem of \emph{text-to-molecule generation} as follows. Our goal is to train a text-to-molecule model $f_\theta$ such that $f_\theta(\mathbf{x})=\mathbf{m}$, where $\mathbf{x}$ is a text description of the desired molecule and $\mathbf{m}$ is the corresponding target molecule (see Table~\ref{fig:chebi_pcdes_visualization} for an example). Recent works \citep{edwards2022translation,pei2023biot5} have shown that such $f_\theta$ can be obtained by fine-tuning a pre-trained language model with description-molecule pairs $\{\mathbf{x}_k, \mathbf{m}_k\}_{k=1}^N$, using the following objective: 
\begin{equation}
\label{eq:training}
    \mathcal{L}(\theta;\mathbf{x}_k, \mathbf{m}_k)\coloneqq 
    \mathcal{L}_{\mathtt{CE}} 
    \Big(
    f_\theta
    (\mathbf{x}_k), \mathbf{m}_k
    \Big),
\end{equation}
where $\mathcal{L}_{\mathtt{CE}}$ denotes cross-entropy loss, and $\mathbf{x}_k$ and $\mathbf{m}_k$ denote the $k$-th text description and the token sequences of the target molecule in the text-to-molecule model's token space, respectively.

Here, the choice of tokenization strategy for $\mathbf{m}_k$ plays a critical role in training an effective $f_\theta$ \citep{pei2023biot5}, as it directly influences how the sequence of tokens captures and represents the structural context of the original molecule. However, previous text-to-molecule models often overlook such importance, relying only on the local connectivity of atoms based on the atom-level tokenization methods, e.g., SMILES \citep{weininger1988smiles} and SELFIES \citep{krenn2020self}. 
In contrast, our contribution resolves the drawbacks of previous tokenization strategies by incorporating substructure-level contextual patterns into the token space of text-to-molecule models. This allows us to represent a molecule in a context-aware manner.

\subsection{\Algname: \algname}
\label{subsec:camt5}

\textbf{Context-aware molecule tokenization.} We propose to construct the molecule token space of \Algname to effectively capture and reflect the structural context of molecules. To achieve this, we consider chemically meaningful fragments, i.e., motifs, as individual tokens in \Algname. This approach differs from previous text-to-molecule models that only rely on atom-level tokens \citep{edwards2022translation,pei2023biot5}. Specifically, we consider the following set of atoms, i.e., a motif, as a single token: (1) atoms forming a ring structure and (2) atoms connected by a non-single bond (see Figure~\ref{fig:concept} for an example). Such groups of atoms are rigidly bound to each other and represent an important structural context, such as resonance \citep{anslyn2006modern}. An atom not associated with (1) and (2) is considered as a single token.

We then propose to represent a molecule as a sequence of motif-level tokens, based on the order of the tree-search algorithm on a tree of motifs. Consider a molecule graph $G=(V,E)$ with the set of atoms $V$ and edges $E$. We construct a tree of motifs $\mathcal{T}(G)=(\mathcal{V}, \mathcal{E})$, where $\mathcal{V}=\{M_i\}_{i=1}^n$ is the set of $n$ motifs with $M_i=(V_i,E_i)$, and $\mathcal{E}$ is the set of bonds between motifs. Here, $\mathcal{T}(G)$ effectively preserves all the information of the original molecule graph $G$, i.e., $V=\cup V_i$ and $E=\cup_i E_i \cup \mathcal{E}$, with context-enriched nodes by replacing atom-level nodes $V$ with motif-level nodes $\mathcal{V}$, satisfying $|\mathcal{V}| \leq |V|$. Consequently, we obtain the sequence of motif tokens by enumerating $\mathcal{V}$ based on the order of the depth-first-search (DFS) algorithm, i.e., $\mathbf{m}_{\text{\Algname}}=[M_1, ..., M_n]$. We then train our text-to-molecule model $f_{\theta}$ with $\{\mathbf{x}_k, \mathbf{m}_{\text{\Algname},k}\}_{k=1}^N$ using the training objective in Eq. (\ref{eq:training}). Note that our method ensures the (1) \emph{validity} of the generated token sequences since we do not introduce tokens that should appear as a pair, c.f., the branch tokens `(' and `)' in SMILES \citep{weininger1988smiles}. Also, our tokens are (2) \emph{non-degenerate} by construction; a single token represents only a single motif, c.f., `$\mathtt{[O]}$' as an oxygen atom or an indicator of a ring system comprising six atoms preceding this token in SELFIES \citep{krenn2020self}. We provide further details of our token space in Appendix~\ref{sup: tokenization_details}.

\input{tables/chebi_main}
\input{tables/pcdes_main}

Our context-enriched tokenization plays a crucial role in discriminating the atoms within different structural contexts. For example, the aromatic carbons in a phenyl group (i.e., $\mathtt{[C][=C][C][=C][C][=C][Ring1][=Branch1]}$ in BioT5; \citealp{pei2023biot5}) and the aliphatic carbons (i.e., $\mathtt{[C][C][C][C][C][C]}$ in BioT5) differ significantly in chemical context, due to resonance and ring structure. However, previous text-to-molecule models do not distinguish the difference between the carbon atoms of each motif, regarding both carbons as the same $\mathtt{[C]}$ token. \Algname resolves this by assigning different tokens for the entire phenyl groups and the carbons in aliphatic carbons.

\noindent\textbf{Importance-based pre-training.} {Previous state-of-the-art text-to-molecule models were pre-trained on vast amounts of tokens from unlabeled molecules \citep{liu2023molxpt,chen2024artificially}. Notably, MolT5 \citep{edwards2021text2mol} and BioT5 \citep{pei2023biot5} demonstrated the effectiveness of the masked language modeling pre-training objective \citep{raffel2020exploring} in enriching the understanding of the molecular domain with unlabeled molecules.}

{In this paper, we advance the masked language modeling \citep{raffel2020exploring} for our motif-level token space, focusing on key motifs during pre-training to better capture molecular structural context. 
To achieve this, we define an importance value $\lambda(M_i)$ for each $M_i \in \mathcal{V}$, reflecting the relative significance of motifs in a given molecule. Based on these pre-defined importance values, we train \Algname with the weighted training loss:}
\begin{align}
\label{eq:confidence}
    \mathcal{L}_{\mathtt{CAMT5}}= \sum_{i=1}^n \lambda(M_i)\mathcal{L}_{\mathtt{MLM}}(M_i),
\end{align}
{where $\mathcal{L}_{\mathtt{MLM}}$ denotes the masked language modeling loss. Here, we find that a simple choice of $\lambda(M_i)$, i.e., the number of atoms in $M_i$, efficiently and effectively improves the generation performance (see Appendix~\ref{sup:experimental_details} for details on the definition of $\lambda$).}

\subsection{Confidence-based ensemble}
\label{subsec:ensemble}
We propose a simple yet effective confidence-based ensemble method to further improve the generation quality of our \Algname. Specifically, we leverage the outputs of other text-to-molecule models, which often use \emph{different} token space, e.g., SMILES \citep{weininger1988smiles} and SELFIES \citep{krenn2020self}. Here, we note that recent ensemble strategies \citep{jiang2024mixtral,sukhbaatar2024branch} only work on the models with the same token space, and thus are not applicable to text-to-molecule models with different tokenizations.

To tackle this issue, we define the \emph{confidence} $C(\mathbf{m}_{i}; f_i, \mathbf{x})$ as the average log-likelihood of the generated tokens, and treat it as a proxy for the quality measure of the generated molecules, i.e.,
\begin{align*}
\nonumber
    C(\mathbf{m}_{i}; f_i, \mathbf{x}) &= \frac{\sum_{j=1}^{K_i}\mathtt{log}P_{f_i}([T_j]|\mathbf{x}, [T_1...,T_{j-1}])}{K_i} \\
    \nonumber
    &= -\mathcal{L}_{\mathtt{CE}}(f_i(\mathbf{x}), \mathbf{m}_{i}),
\end{align*}
where $f_i$ is the $i$-th text-to-molecule model and $\mathbf{m}_{i} = [T_1,...,T_{K_i}]$ be the generated $K_i$ tokens from $f_i$ to the given description $\mathbf{x}$. Then, we define the confidence-based ensemble $f_\mathtt{ens}$ with $f_1,...,f_n$ as follows:
\begin{align}
\label{eq:ensemble}
f_\mathtt{ens}(\mathbf{x}) &= \mathbf{m}_k\text{, where }k = \mathtt{argmax}_{i}\text{ }C(\mathbf{m}_{i}; f_i, \mathbf{x}).
\end{align}

We note that this ensemble strategy is particularly useful in practical scenarios. Previously, people simply chose the best-performing model among the existing text-to-molecule models, ignoring other on-average under-performing models. However, when the selected model is not \emph{confident} in a certain text description, other models may provide more confident alternatives. In this case, our confidence-based ensemble strategy can be applied to further improve the performance of the best-performing model, i.e., \Algname, with the help of other existing models, i.e., MolT5 and BioT5.

%% file: tables/chebi_main.tex
\begin{table*}[t]

\small
\begin{center}
\resizebox{1.0\textwidth}{!}{%

\begin{tabular}{l|c|c|c|cccccc}
\toprule

 \textbf{Model} & \textbf{\#Params.} & \textbf{Representation} &\textbf{Train Tokens} & \textbf{Exact $\uparrow$} & \textbf{MACCS $\uparrow$} & \textbf{RDK $\uparrow$} & \textbf{Morgan $\uparrow$} & \textbf{Valid. $\uparrow$} \\ \midrule
 $\text{RNN}$ & 56M & SMILES & - & 0.005 & 0.591 & 0.400 &  0.362 & 0.542 \\
 $\text{Transformer}$ & 76M & SMILES & - & 0.000 &  0.480 & 0.320 & 0.217 & 0.906\\ \midrule
 $\text{T5}_\mathtt{small}$ & 77M & SMILES & - & 0.064 &  0.704 & 0.578 & 0.525 & 0.608 \\ 
 $\text{T5}_\mathtt{base}$ & 248M & SMILES & - & 0.069 &  0.731 & 0.605 & 0.545 & 0.660 \\ 
 $\text{T5}_\mathtt{large}$ & 783M & SMILES & - & 0.279 &  0.823 & 0.731 & 0.670 & 0.902 \\ 
 
 \midrule
 $\text{MolT5}_\mathtt{small}$ & 77M & SMILES  & 66B & 0.079 &  0.703 & 0.568 & 0.517 & 0.721 \\ 
 $\text{MolT5}_\mathtt{base}$ & 248M & SMILES  & 66B & 0.081 &  0.721 & 0.588 & 0.529 & 0.772 \\ 
 $\text{MolT5}_\mathtt{large}$ & 783M & SMILES  & 66B & 0.311 &  0.834 & 0.746 & 0.684 & 0.905 \\ 
 \midrule

 $\text{GPT-3.5-turbo}$ & $>$175B & SMILES & - & 0.019 &  0.705 & 0.462 & 0.367 & 0.802 \\ 
 MolReGPT & $>$175B & SMILES & - & 0.139 &  0.847 & 0.708 & 0.624 & 0.887 \\ 
 \midrule

 $\text{MolXPT}$ & 350M & SMILES & 1.8B & 0.215 &  0.859 & 0.757 & 0.667 & 0.983 \\ 
 \midrule

 $\text{BioT5}_\mathtt{base}^*$ & 252M & SELFIES & 69B & 0.413 &  \textbf{0.886} & 0.801 & 0.734 & \textbf{1.000} \\  \midrule
$\text{MolT5}_\mathtt{base}^\dagger$ & 248M & SMILES &  1.6B & 0.326 & 0.847 & 0.797 & 0.720 & 0.950 \\ 
$\text{BioT5}_\mathtt{base}^\dagger$ & 252M & SELFIES & 1.6B & 0.344 & 0.842 & 0.773 & 0.664 & \textbf{1.000} \\ 
 
 \midrule

 \rowcolor{tablegreen}$\textbf{CAMT5}_\mathtt{small}$ \textbf{(Ours)}& 103M & \textbf{Motif (Ours)} & 1.6B & 0.391 & 0.874 & 0.827 & 0.727 & \textbf{1.000} \\ 
 \rowcolor{tablegreen}$\textbf{CAMT5}_\mathtt{base}$  \,\,\textbf{(Ours)} & 286M & \textbf{Motif (Ours)} & 1.6B & {0.422} &  0.882 & {0.834} & {0.742} & \textbf{1.000} \\ 
 \rowcolor{tablegreen}$\textbf{CAMT5}_\mathtt{large}$ \textbf{(Ours)}& 836M & \textbf{Motif (Ours)} & 1.6B & \textbf{0.430} & 0.885 & \textbf{0.840} & \textbf{0.749} & \textbf{1.000} \\ 
\bottomrule

\end{tabular}
}
\end{center}
\vspace{-0.15in}
\caption{Quantitative results of the text-to-molecule generation task in the CheBI-20 \citep{edwards2021text2mol} benchmark. $\mathtt{small}$, $\mathtt{base}$ and $\mathtt{large}$ denote that the model is derived from the T5-small, T5-base and T5-large \citep{raffel2020exploring}, respectively. $\#$Params denotes the number of parameters in each text-to-molecule model. Train Tokens refers to the number of molecule-related pre-training tokens. $*$ denotes that the model is pre-trained with an additional non-public high-quality molecular dataset, which is not available for us. $\dagger$ denotes that the model is trained with the same training configuration, e.g., training dataset, as ours. We highlight the best score in bold. 
}
\label{tab:chebi_main}
\vspace{-0.1in}
\end{table*}

%% file: tables/pcdes_main.tex
\begin{table*}[t]

\small
\begin{center}
\begin{tabular}{l|c|c|c|cccccc}
\toprule
\textbf{Model} & \textbf{\#Params.} & \textbf{Representation} & \textbf{Exact $\uparrow$} & \textbf{MACCS $\uparrow$} & \textbf{RDK $\uparrow$} & \textbf{Morgan $\uparrow$} & \textbf{Valid. $\uparrow$} \\ \midrule
$\text{MolT5}_\mathtt{base}^\dagger$ & 248M & SMILES  & 0.151 & 0.578 & 0.523 & 0.417 & 0.793\\ 
$\text{BioT5}_\mathtt{base}^\dagger$ & 252M & SELFIES & 0.132 & 0.695 & 0.624 & 0.458 & \textbf{1.000} \\ \midrule
\rowcolor{tablegreen}\textbf{$\text{CAMT5}_\mathtt{base}$ (Ours)} & 286M & \textbf{Motif (Ours)} & \textbf{0.196} & \textbf{0.738} & \textbf{0.679} & \textbf{0.528} & \textbf{1.000} \\
\bottomrule
\end{tabular}
\end{center}
\vspace{-0.15in}
\caption{Quantitative results of the text-to-molecule generation task in PCDes \citep{zeng2022deep}. $\dagger$ denotes that the model is trained with the same training configuration, e.g., training dataset, as ours. We bold the best score.}

\label{tab:pcdes_main}
\vspace{-0.1in}
\end{table*}

%% file: 4_experiments.tex
\input{figures/ensemble_visualization}

\input{tables/ensemble_main.tex}

\section{Experiments}
\label{sec:experiments}

We verify the effectiveness of \Algname through extensive experiments. In Section~\ref{subsec:experimental_setup}, we explain our experimental setups. Section~\ref{subsec:main_experiment} presents the text-to-molecule generation results on the ChEBI-20 and PCDes benchmarks. In Section~\ref{subsec:ensemble}, we present the results of our confidence-based ensemble strategy. In Section~\ref{subsec:applications}, we show the text-conditional molecule modification task results. In Section~\ref{subsec:analysis}, we provide ablation studies on components of \Algname. We provide additional experimental results and analyses in Appendix~\ref{sup:additional_experiments}.

\subsection{Experimental setup}
\label{subsec:experimental_setup}

\textbf{Baselines.} We consider the recently proposed state-of-the-art text-to-molecule models: MolT5 \citep{edwards2022translation}, MolReGPT \citep{li2024empowering}, MolXPT \citep{liu2023molxpt}, and BioT5 \citep{pei2023biot5}. These models are based on atom-wise tokenization, i.e., SMILES and SELFIES.

\noindent\textbf{Datasets.} We evaluate the text-to-molecule generation performance of text-to-molecule models on two popular benchmarks, ChEBI-20 \citep{edwards2021text2mol} and PCDes \citep{zeng2022deep}. In addition, we construct a new dataset of 34k description-molecule pairs from the PubChem database, ensuring no overlap with the molecules in ChEBI-20 or PCDes. This dataset is used to train our \Algname, as well as $\dagger$-marked MolT5 and BioT5 models (see Table~\ref{tab:chebi_main}). Further details are provided in Appendix~\ref{sup:dataset_details}.

\input{figures/case_study}

\noindent\textbf{Training setup.} Following the previous practices \citep{edwards2021text2mol, pei2023biot5}, we pre-train text-to-molecule models with publically available uni-modal datasets, i.e., C4 \citep{raffel2020exploring} for the text corpus and ZINC-15 \citep{sterling2015zinc} for the molecule corpus. We note that the previous models, e.g., MolT5 and BioT5, are trained with different datasets, which limits a genuine comparison of proposed methods. For example, the official BioT5 model benefits from an additional non-public pre-training dataset. To alleviate this issue, we have aligned the pre-training and fine-tuning configurations of each model and marked $\dagger$, e.g., as shown in Table~\ref{tab:chebi_main}. We provide further details of experimental setups in Appendix~\ref{sup:experimental_details}.

\noindent\textbf{Metrics.} For an extensive evaluation of text-to-molecule generation, we utilize various metrics that reflect the quality of the generated molecules. The detailed description of metrics are as follows:
\begin{itemize}[topsep=1.5pt,itemsep=1.0pt,leftmargin=5.5mm]
    \item [$\bullet$] \textbf{Exact}: The ratio of the generated molecules that exactly match with the target molecule.
    \item [$\bullet$] \textbf{MACCS/RDK/Morgan Fingerprint Tanimoto Similarity (MACCS/RDK/Morgan)}: Metrics that measure the fingerprint-level similarity between the generated molecule and the target molecule. MACCS \citep{durant2002reoptimization}, RDK \citep{schneider2015get}, and Morgan \citep{rogers2010extended} fingerprints are used. If the generated token sequence is not a valid molecule, we set this score as 0. 
    For each dataset, we report the average scores of the generated molecules in each dataset. 
    \item [$\bullet$] \textbf{Validity (Valid.)}: The ratio of the generated token sequences that are valid molecules.\footnote{For BioT5 and \Algname, Validity is guaranteed to be 1.0 due to the characteristics of used token representations.}
\end{itemize}

\input{tables/molecule_modification}

\subsection{Main experiments}
\label{subsec:main_experiment}
Table~\ref{tab:chebi_main} and~\ref{tab:pcdes_main} summarize the quantitative results of the text-to-molecule generation tasks in ChEBI-20 \citep{edwards2021text2mol} and PCDes \citep{zeng2022deep}, respectively. In both benchmarks, our \Algname consistently outperforms the baseline text-to-molecule models by generating desirable molecules corresponding to the text description. 

\noindent\textbf{Results on ChEBI-20.} In the ChEBI-20 benchmark \citep{edwards2021text2mol}, $\text{\Algname}$ highly outperforms the state-of-the-art text-to-molecule model, BioT5 \citep{pei2023biot5}, which leverages an additional non-public high-quality pre-training dataset. For example, \Algname shows superior performance in generating molecules that exactly match the given text descriptions, improving the Exact score by 0.413 $\rightarrow$ 0.430. Furthermore, \Algname generates molecules more similar to the given description, achieving higher fingerprint similarity-based scores, e.g., 0.801 $\rightarrow$ 0.840 and 0.734 $\rightarrow$ 0.749 in the RDK and Morgan similarity scores, respectively. Notably, \Algname achieves these improvements with only 2\% of molecule-related pre-training tokens compared to BioT5, underscoring the superiority of our molecule tokenization and importance-based pre-training strategy.

For an extensive comparison with baselines in a fair setup, we also provide the results under the same training datasets and configurations as our \Algname (denoted by MolT5$_{\mathtt{base}}^\dagger$ and BioT5$_{\mathtt{base}}^\dagger$). Within this setup, our model of a similar size, i.e., $\text{\Algname}_{\mathtt{base}}$, demonstrates a significant performance improvement, achieving the Exact score by 0.344 $\rightarrow$ 0.422. Moreover, it is noteworthy that even with the smaller variant, i.e., $\text{\Algname}_{\mathtt{small}}$, our method consistently outperforms both MolT5$^\dagger_{\mathtt{base}}$ and BioT5$^\dagger_{\mathtt{base}}$ across all evaluated metrics.
These results underscore \Algname's strong efficacy in generating desired molecules and establish it as a promising approach for text-to-molecule generation tasks.

\noindent\textbf{Results on PCDes.} Table~\ref{tab:pcdes_main} shows that \Algname is also effective in the more challenging PCDes \citep{zeng2022deep} benchmark, with improvements such as 0.151 $\rightarrow$ 0.196 in the Exact score and 0.624 $\rightarrow$ 0.679 in the RDK score. This highlights the robustness and applicability of our \Algname across various text-to-molecule generation tasks.

\input{tables/importance_ablation}

\subsection{Confidence-based ensemble}
\label{subsec:results_on_ensemble}
In Table~\ref{tab:ensemble_main}, we report the quantitative results of the selected molecules from our confidence-based ensemble strategy (see Eq.(\ref{eq:ensemble})). In this experiment, we construct an ensemble model based on the state-of-the-art text-to-molecule models, e.g., MolT5 \citep{edwards2022translation}, BioT5 \citep{pei2023biot5}, and our \Algname. During ensemble, we make sure that the generated molecules are all valid, by ignoring the output from MolT5 when it does not correspond to a valid molecule. We note that this does not incur an additional computational overhead, since verifying the validity of the generated output does not require computational cost. Overall, {our ensemble} strategy significantly improves the performance of existing text-to-molecule models, e.g., 0.430 $\rightarrow$ 0.472 and 0.196 $\rightarrow$ 0.213 in the Exact score on the CheBI-20 and PCDes benchmarks, respectively. In Table~\ref{fig:ensemble_visualization}, we provide some examples where our \Algname is not quite confident in its output, and other models, i.e., MolT5 and BioT5, generate more confident molecules. In these cases, the ensemble strategy selects the molecules generated by MolT5 or BioT5, which are indeed more similar to the target molecules. In summary, our ensemble strategy effectively leverages on-average underperforming models, i.e., MolT5 and BioT5, to further improve the output of the best-performing model, i.e., \Algname, through our carefully designed confidence-based ensemble strategy.

\subsection{Text-conditional molecule modification}
\label{subsec:applications}

In this section, we verify the potential of our \Algname in the context of \emph{modifying} molecules based on additional text prompt conditions.
To achieve molecule modification, we consider a text-to-molecule model $f$, where $f(\mathbf{x}) = \mathbf{m}$ maps a molecule description $\mathbf{x}$ to its corresponding molecule $\mathbf{m}$. Then, we slightly alter the description $\mathbf{x}$ by appending an additional condition prompt, such as $\mathbf{x}' = \mathbf{x} + \text{``}\mathsf{Make\,\,it\,\,\emph{insoluble}\,\,in\,\,water}$''. The resulting modified molecule $\mathbf{m}' = f(\mathbf{x}')$ is expected to (1) maintain structural similarity to the original molecule $\mathbf{m}$ and (2) faithfully capture the additional conditional text in $\mathbf{x}'$. Although previous studies have considered the modification of molecules based on numerical value conditions \citep{chen2021deep, zhu2024sample}, the exploration of molecule modification conditioned on textual descriptions remains relatively under-explored \citep{zhu20243m}, despite its practical potential.

In Table~\ref{fig:case_study}, we consider the descriptions in the ChEBI-20 test set where $\text{MolT5}$, $\text{BioT5}$, and $\text{\Algname}$ each generate the same molecule represented in the Query column. We then generate molecules with additional condition prompt in addition to the original description. 
The results show that our \Algname demonstrates meaningful modification capabilities by excelling in two key aspects: (1) preserving the critical substructures of the original molecule in the Query column, such as the {N-acyl group}, and (2) effectively incorporating additional prompts, as evidenced by the resulting LogP values. We hypothesize that this improvement stems from our unique motif-level tokenization strategy, which is advantageous for incorporating motifs closely related to molecular properties.

In Table~\ref{tab:molecule_modification}, we compare the performance of each model based on (i) the MACCS similarity to the original target molecule, and (ii) $\Delta$LogP, defined as the difference in LogP values between the generated molecule and the target molecule. The results show that \Algname generates molecules that are structurally closer to the target molecule, while also reflecting the intended property condition.

\subsection{Ablation studies}
\label{subsec:analysis}
In this section, we verify the effectiveness of the core components of \Algname, context-aware tokenization and importance-based pre-training strategy. As demonstrated in Table~\ref{tab:analysis} and Figure~\ref{fig:ablation_token_number}, our \Algname without importance-based training (i.e., the third row of the table) already improves the previous best-performing models, e.g., MolT5 \citep{edwards2022translation} and BioT5 \citep{pei2023biot5}. In other words, this result shows the superiority of our tokenization compared to the previous atom-wise tokenizations, such as SMILES \citep{weininger1988smiles} in MolT5 and SELFIES \citep{krenn2020self} in BioT5. Furthermore, \Algname with the importance-based pre-training strategy (i.e., the last row of the table) significantly outperforms the model pre-trained with the conventional masked language modeling \citep{raffel2020exploring} objective (i.e., the third row of the table). This result underscores that guiding the text-to-molecule model to focus more on key substructures is largely advantageous in learning the text-conditional molecule distribution. Overall, these results demonstrate that our carefully designed context-aware tokenization and the importance-based pre-training strategy play crucial roles in understanding molecules, and thus improving text-to-molecule generation performance.

%% file: figures/ensemble_visualization.tex
\begin{table*}[ht]

\begin{center}
\resizebox{1.0\textwidth}{!}{
\small
\begin{tabular}{c|ccc|c|c}
\toprule
 Description &  $\text{MolT5}$ & 
$\text{BioT5}$  &
  \textbf{$\text{\Algname}$ (Ours)} & \textbf{Ensemble (Ours)} & Target
\\ \midrule
\includegraphics[height=1.0in,valign=c]{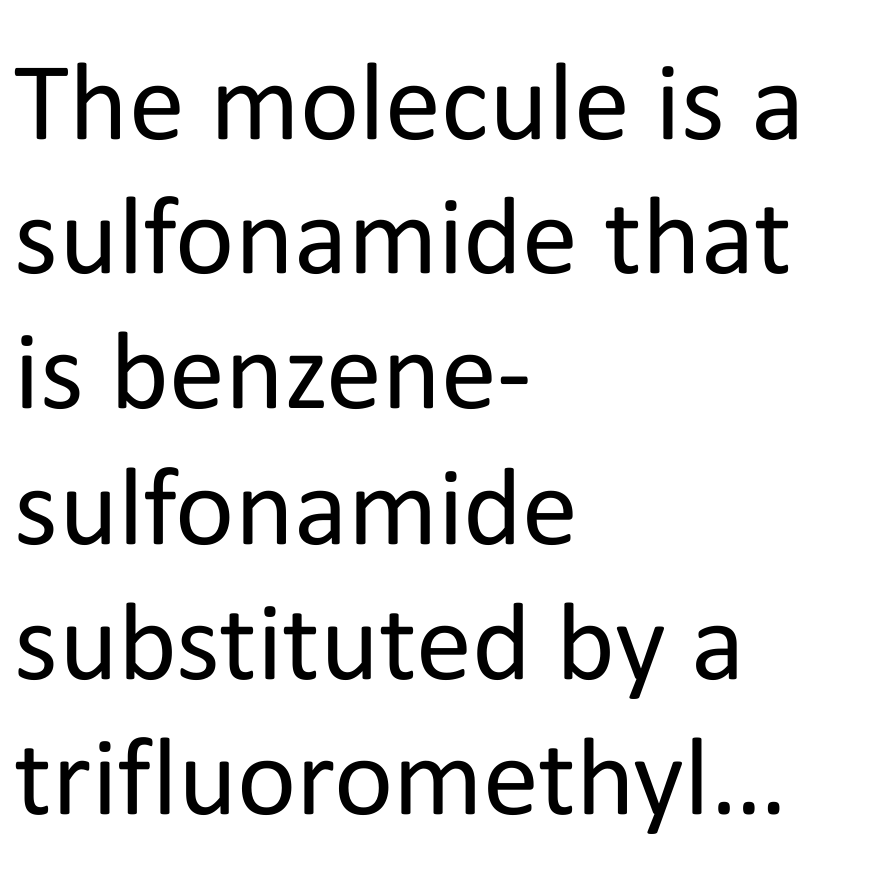}  &
\includegraphics[height=1.0in,valign=c]{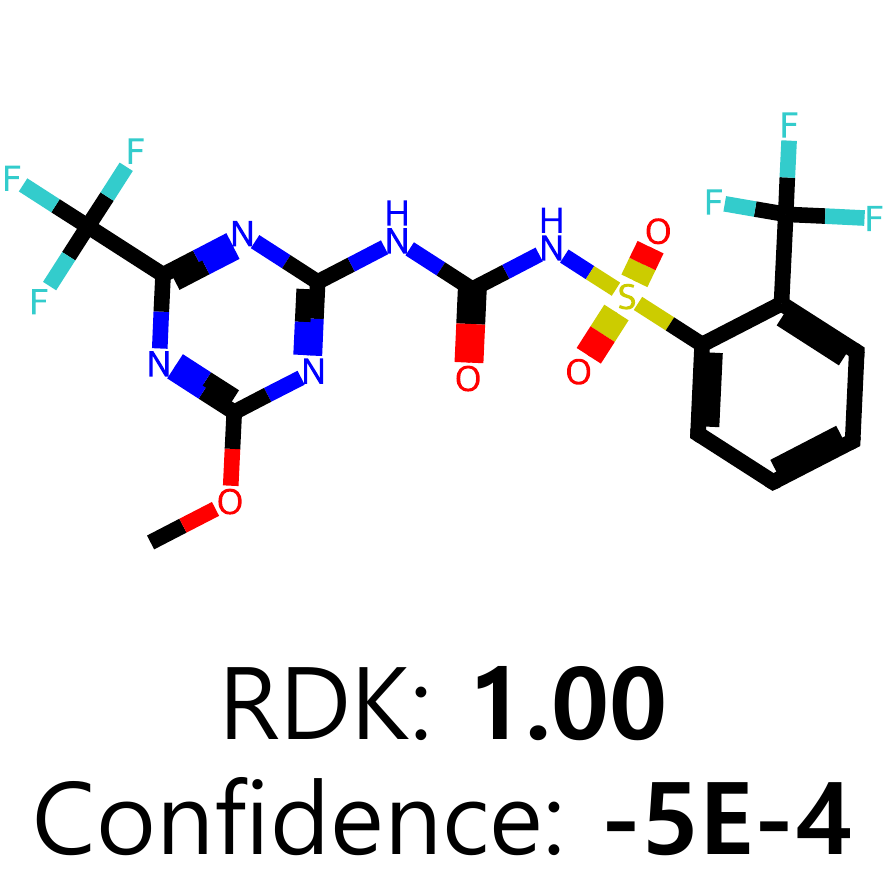} & 
\includegraphics[height=1.0in,valign=c]{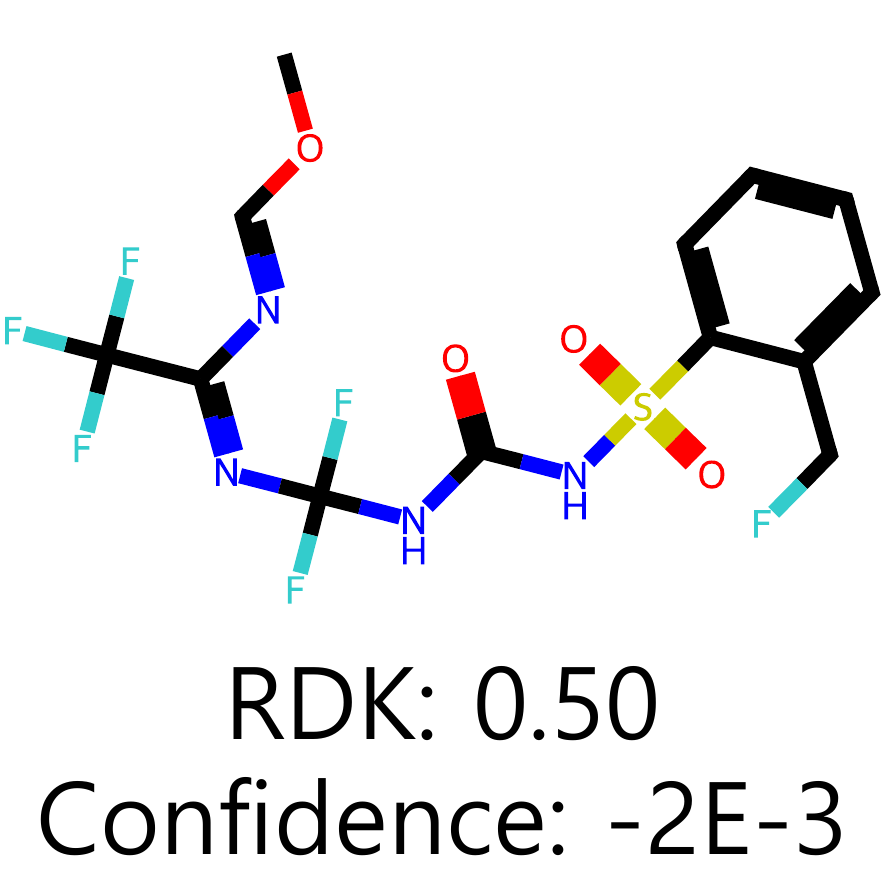}   &  
\includegraphics[height=1.0in,valign=c]{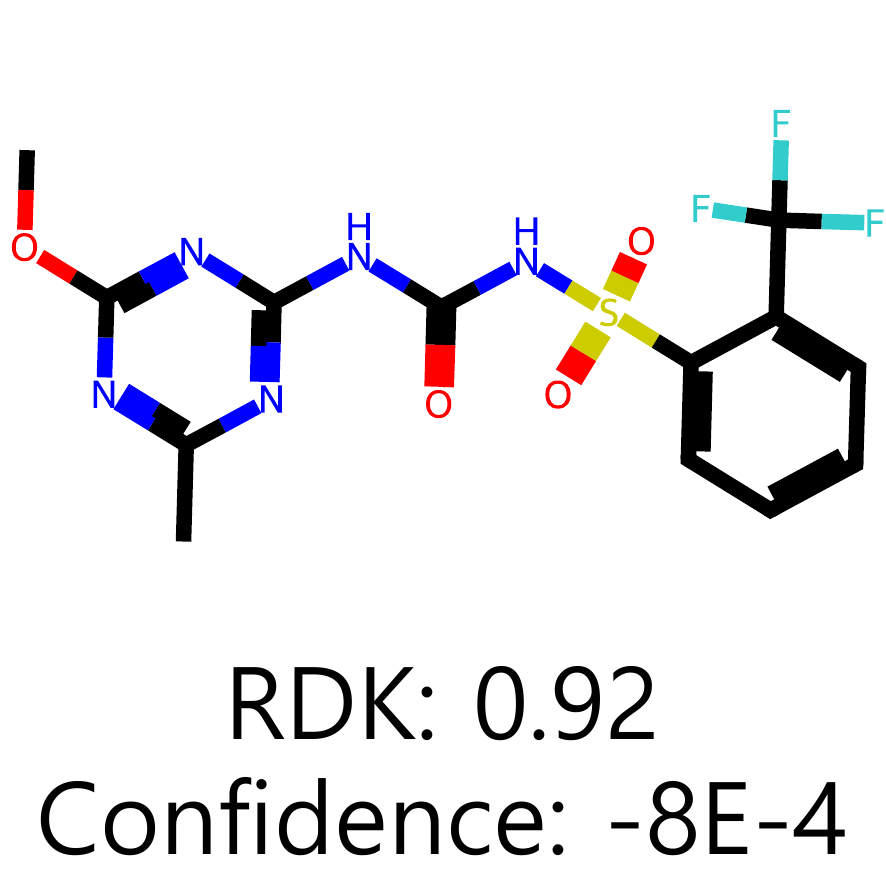} & \includegraphics[height=1.0in,valign=c]{figures/chebi-ensemble-molt5.pdf}  &
\includegraphics[height=1.0in,valign=c]{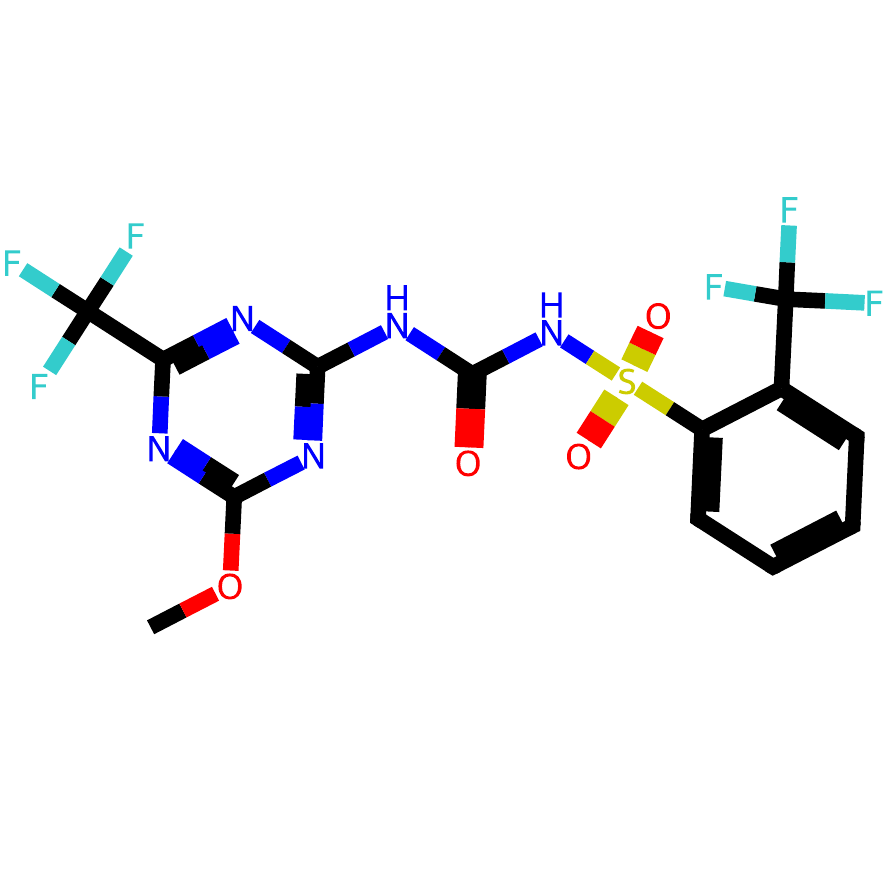}   \\\midrule
 \includegraphics[height=1.0in,valign=c]{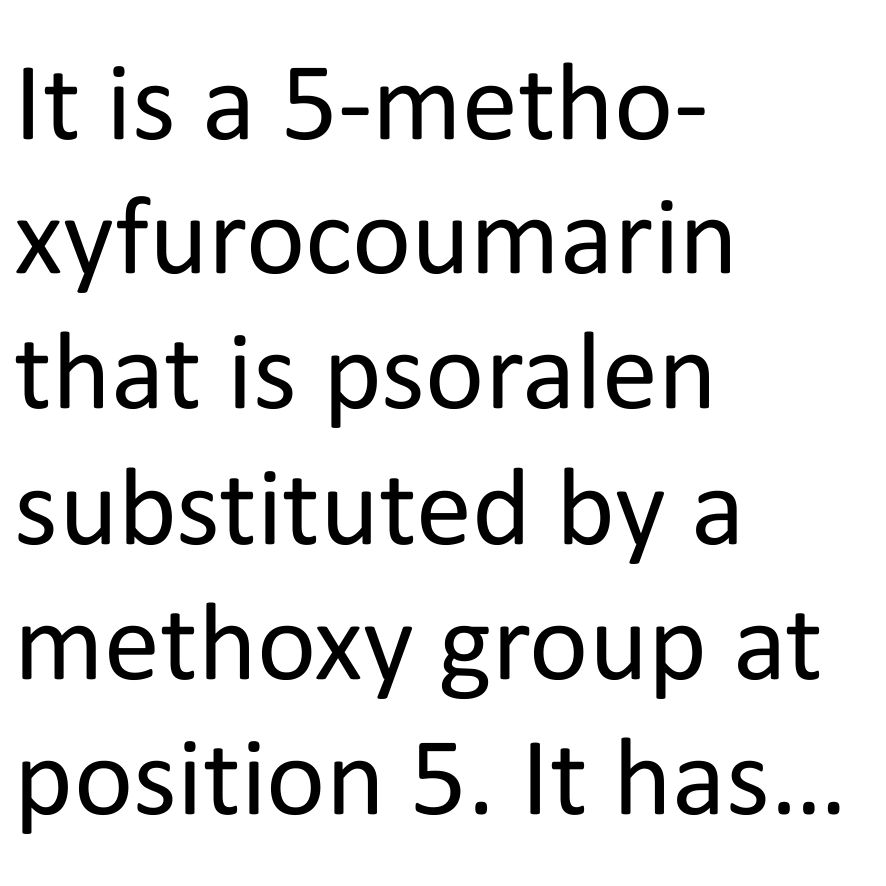} &   
\includegraphics[height=1.0in,valign=c]{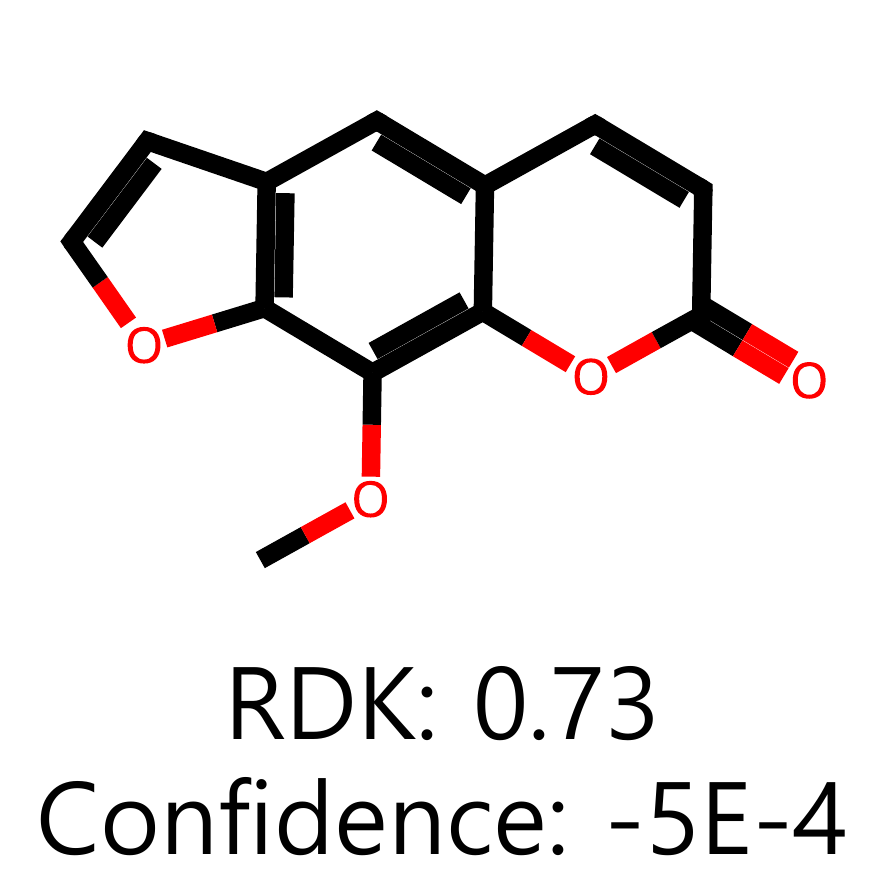} & 
\includegraphics[height=1.0in,valign=c]{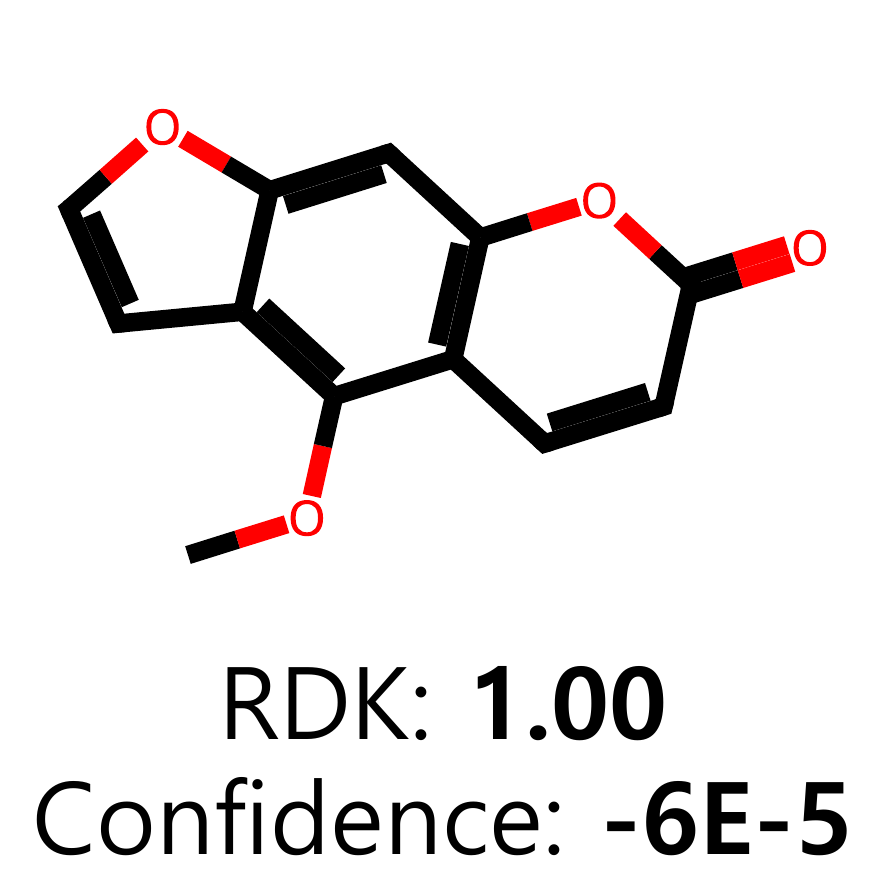} & 
\includegraphics[height=1.0in,valign=c]{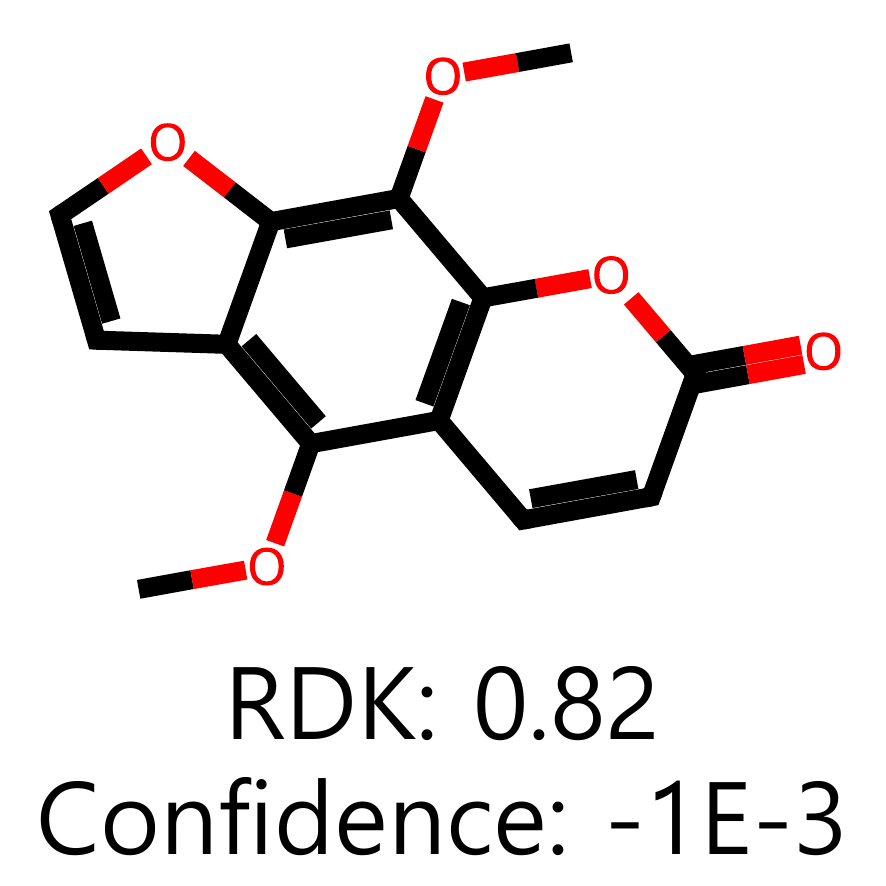} 
& 
\includegraphics[height=1.0in,valign=c]{figures/pcdes-ensemble-biot5.pdf}
&
\includegraphics[height=1.0in,valign=c]{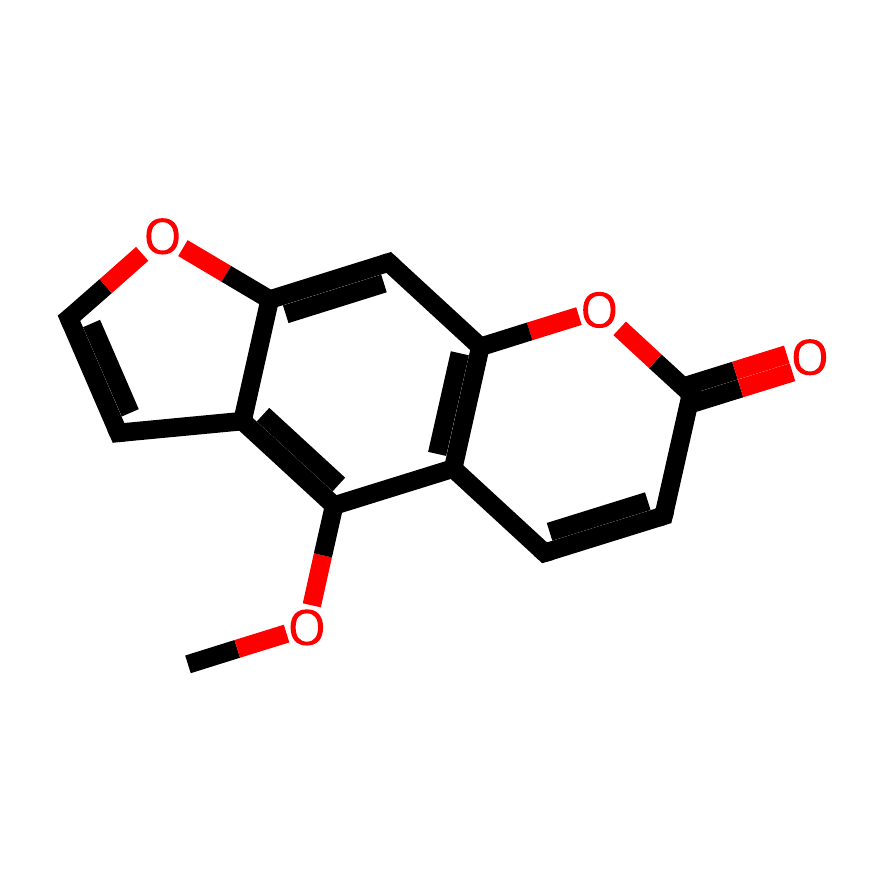} \\
\bottomrule
\end{tabular}
}
\end{center}
\vspace{-0.15in}
\caption{
Visualizations of the confidence-based ensemble on CheBI-20 (\citealt{edwards2021text2mol}; the first row) and PCDes (\citealt{zeng2022deep}; the second row). We visualize the cases that other models, i.e., MolT5 and BioT5, {help} our \Algname through ensemble when the confidence (maximally 0.00) of our generated molecule is relatively low. We report the confidences and the RDK scores below each visualization. Our ensemble strategy selects the molecule with the highest confidence as the output of ensemble (see Eq. (\ref{eq:ensemble})). We bold the highest score.
}

\label{fig:ensemble_visualization}

\end{table*}

%% file: tables/ensemble_main.tex
\begin{table}[ht]

\begin{center}
\resizebox{1.0\linewidth}{!}{%
\begin{tabular}{l|ccccc}
\toprule

\textbf{Model} & \textbf{Exact} $\uparrow$ & \textbf{MACCS} $\uparrow$ & \textbf{RDK} $\uparrow$ & \textbf{Morgan} $\uparrow$ & \textbf{Valid.} $\uparrow$ \\ \midrule
\rowcolor{Gray} \multicolumn{6}{c}{Results on the CheBI-20 benchmark.}\\\midrule
{$\text{MolT5}$} & 0.326 & 0.847 & 0.797 & 0.720 & 0.950 \\ 
{$\text{BioT5}$} & {0.413} & {0.886} & {0.801} & {0.734} & \textbf{{1.000}} \\ 
{\textbf{$\text{\Algname}$}} &{0.430} & {0.885} & {0.840} & {0.749} & \textbf{{1.000}} \\ \midrule
\rowcolor{tablegreen}\textbf{Ensemble} & \textbf{{0.472}} & \textbf{{0.902}} & \textbf{{0.860}} & \textbf{{0.781}} & \textbf{{1.000}} \\ 
 \midrule

\rowcolor{Gray}\multicolumn{6}{c}{Results on the PCDes benchmark.}\\\midrule
{$\text{MolT5}$} & 0.151 & 0.578 & 0.523 & 0.417 & 0.793 \\ 
{$\text{BioT5}$} & {0.132} & {0.695} & {0.624} & {0.458} & \textbf{{1.000}} \\ 
{\textbf{$\text{\Algname}$}} &{0.196} & {0.738} & {0.679} & {0.528} & \textbf{{1.000}} \\  \midrule
\rowcolor{tablegreen}\textbf{Ensemble} & \textbf{{0.213}} & \textbf{{0.755}} & \textbf{{0.695}} & \textbf{{0.554}} & \textbf{{1.000}}  \\  

\bottomrule

\end{tabular}
}
\end{center}
\vspace{-0.15in}
\caption{Quantitative results of our confidence-based ensemble on the CheBI-20 \citep{edwards2021text2mol} and PCDes \citep{zeng2022deep} benchmarks. We report the ensemble results based on the best-performing models of $\text{MolT5, BioT5, \Algname}$ in Table~\ref{tab:chebi_main} and~\ref{tab:pcdes_main}, respectively. We highlight the best score in bold.}
\label{tab:ensemble_main}

\end{table}

%% file: figures/case_study.tex
\begin{table*}[t]
\begin{center}
\small
\begin{tabular}{c|cc|cc|cc}
\toprule
 Query & \multicolumn{2}{c|}{$\text{MolT5}$} & \multicolumn{2}{c|}{$\text{BioT5}$} & \multicolumn{2}{c}{\textbf{$\text{\Algname}$ (Ours)}}
\\ \midrule
\rowcolor{Gray} \multicolumn{7}{c}{Prompt: $``\mathsf{The\,\,molecule\,\,is\,\,an\,\,N\text{-}acyl\,\,acid\,\,ester\,\,...\,\,Make\,\,it\,\,\emph{soluble}\,\,in\,\,water.}$'' (Lower LogP is better)}\\\midrule
\includegraphics[height=0.7in,valign=c]{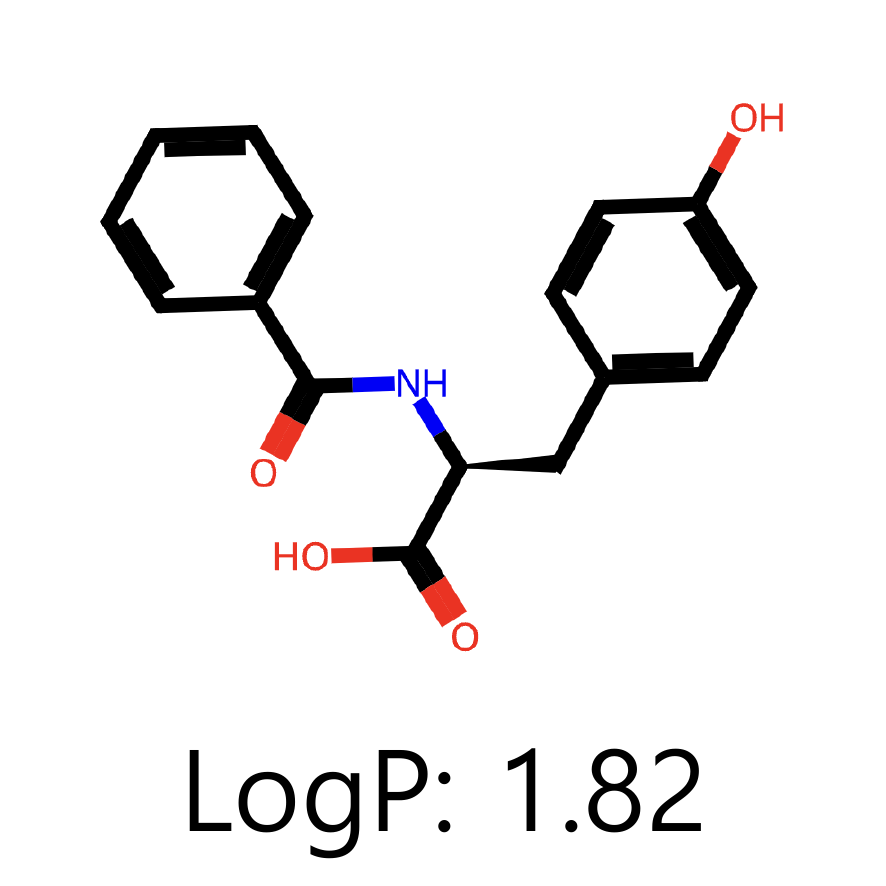}     &
\includegraphics[height=0.7in,valign=c]{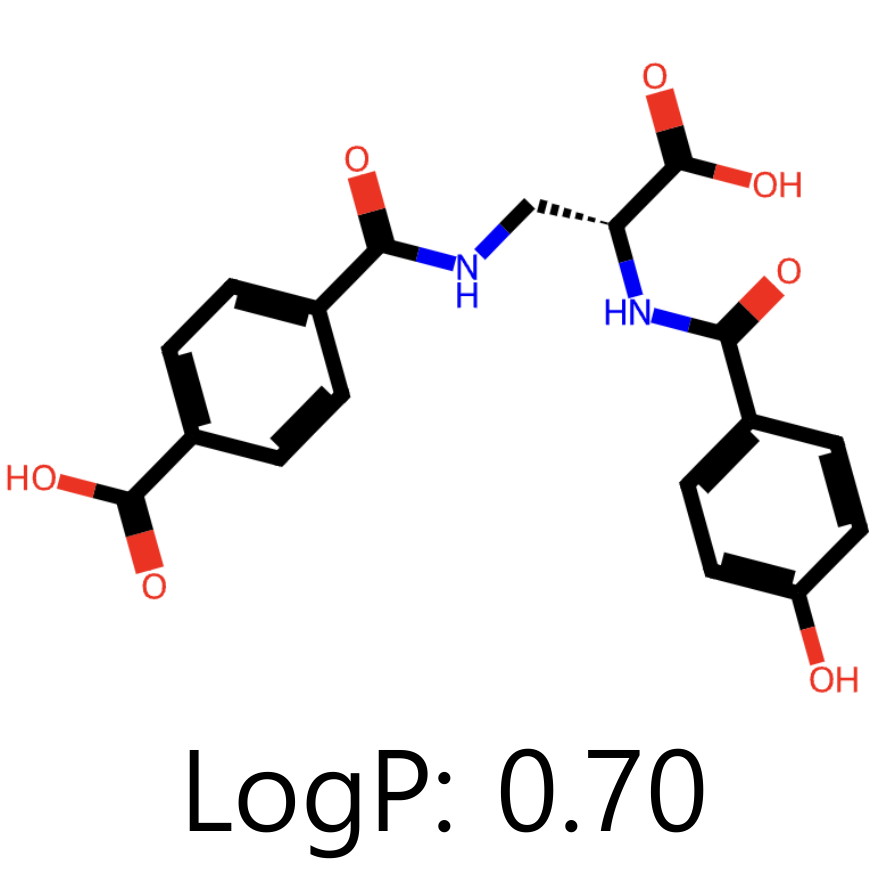} & 
\includegraphics[height=0.7in,valign=c]{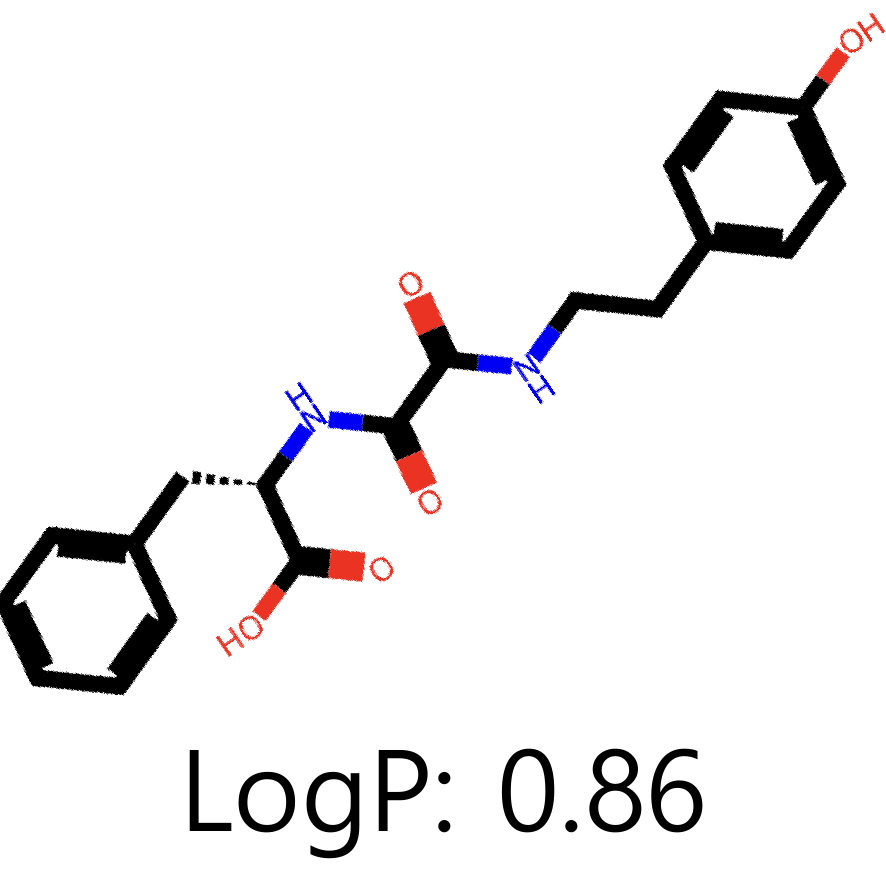}   &  
\includegraphics[height=0.7in,valign=c]{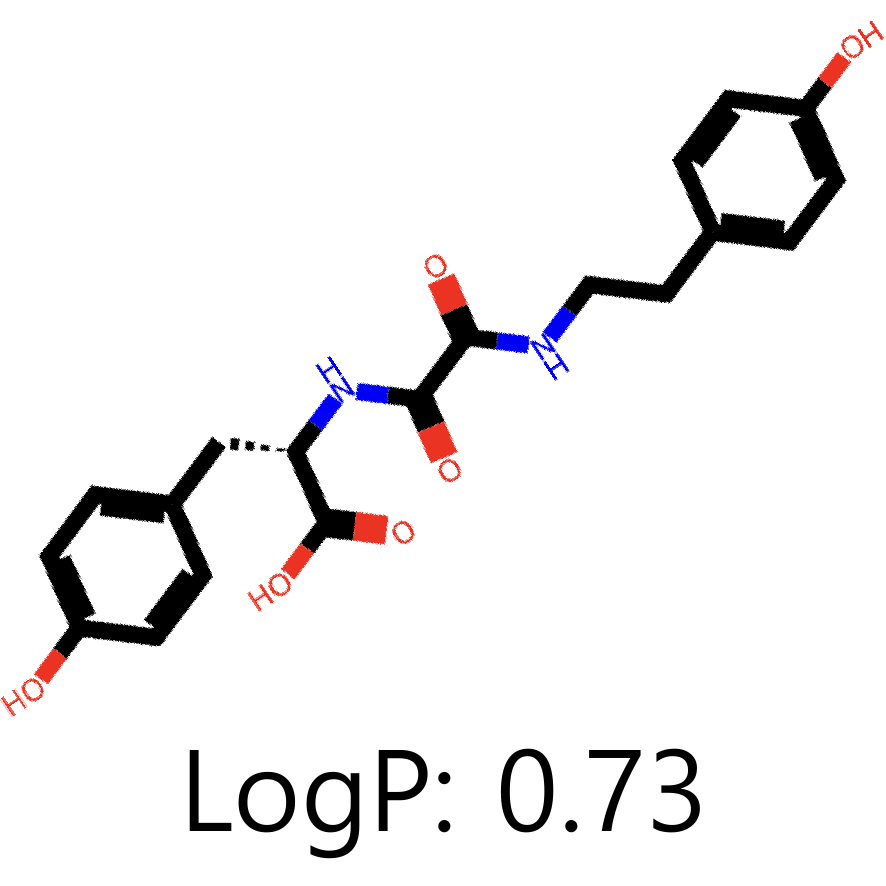} & 
\includegraphics[height=0.7in,valign=c]{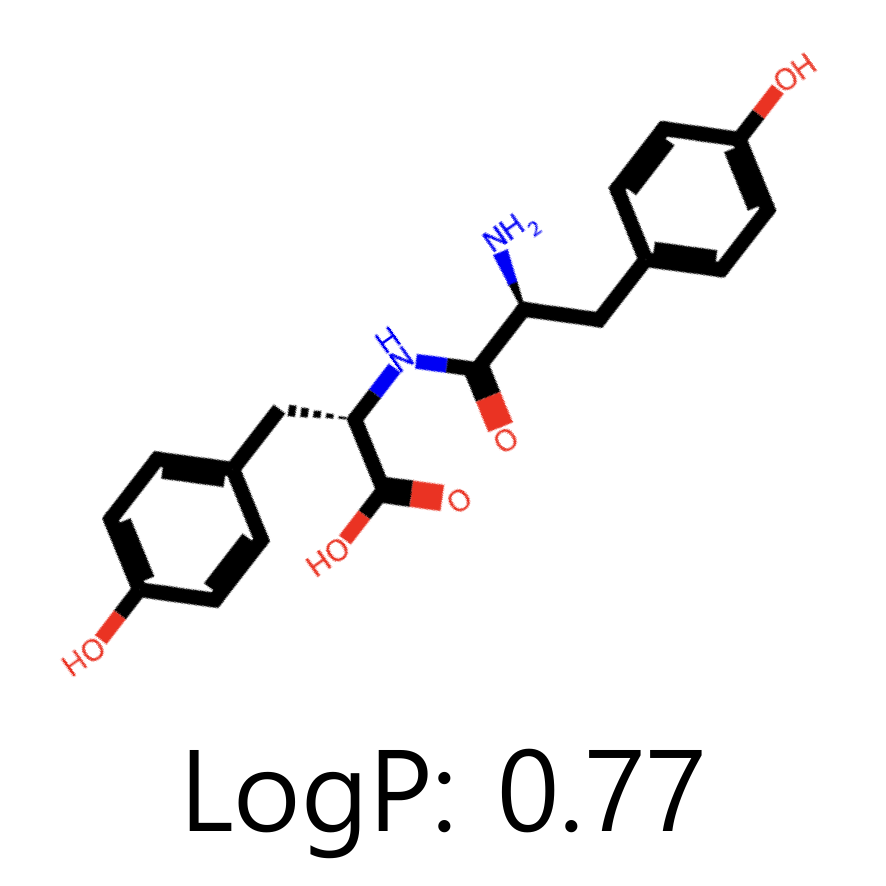}  & \includegraphics[height=0.7in,valign=c]{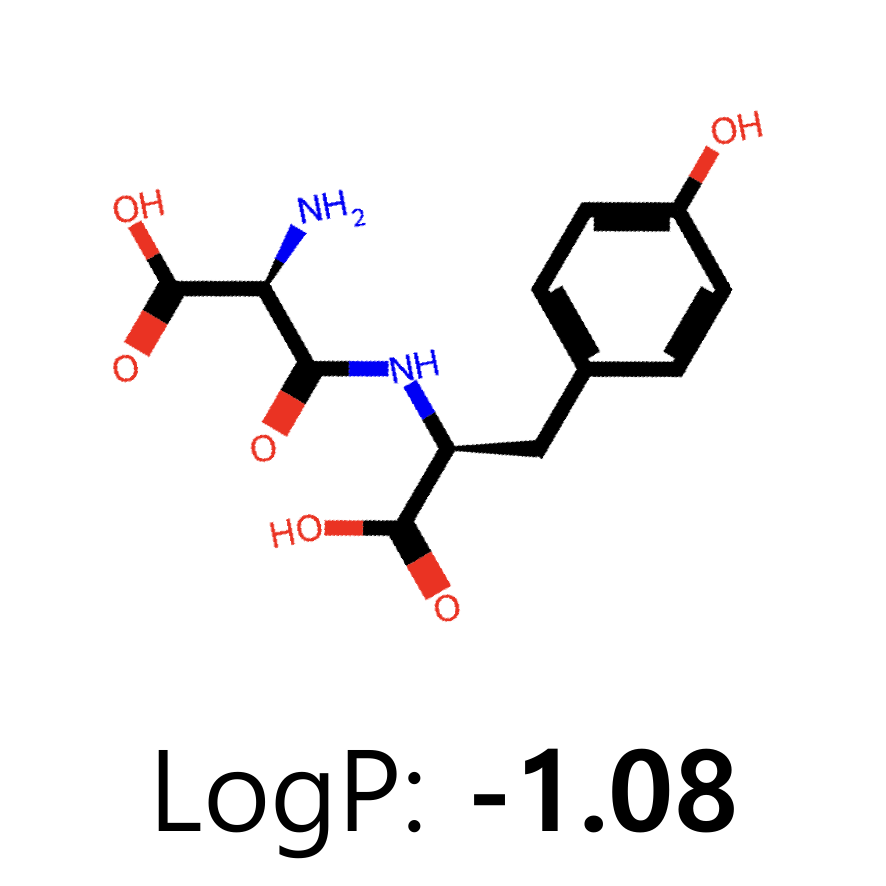}  & \includegraphics[height=0.7in,valign=c]{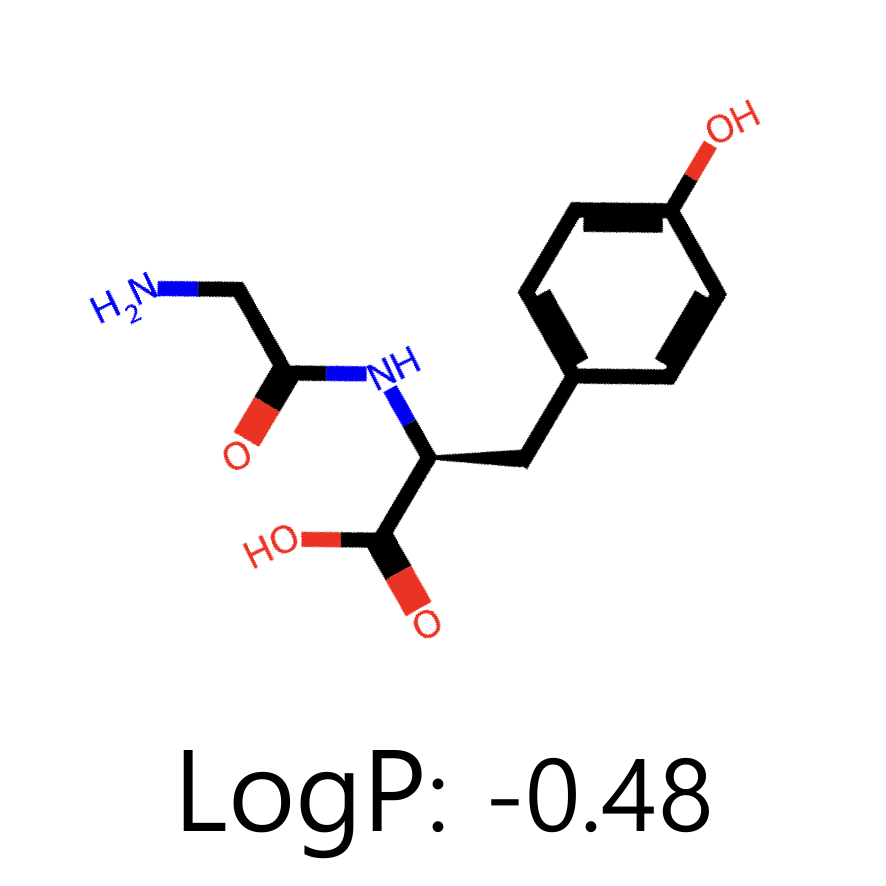}   \\\midrule
\rowcolor{Gray} \multicolumn{7}{c}{Prompt: $``\mathsf{The\,\,molecule\,\,is\,\,a\,\,member\,\,of\,\,ureas...\,\,Make\,\,it\,\,\emph{insoluble}\,\,in\,\,water.}$'' (Higher LogP is better)}\\\midrule

 \includegraphics[height=0.7in,valign=c]{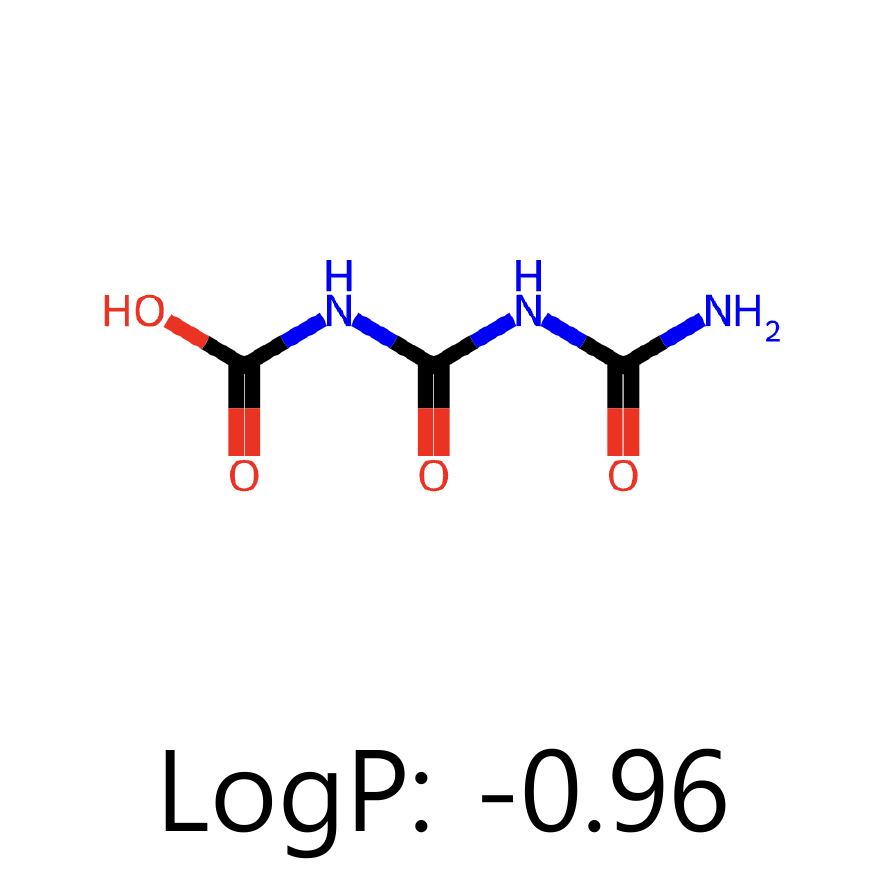}     & \includegraphics[height=0.7in,valign=c]{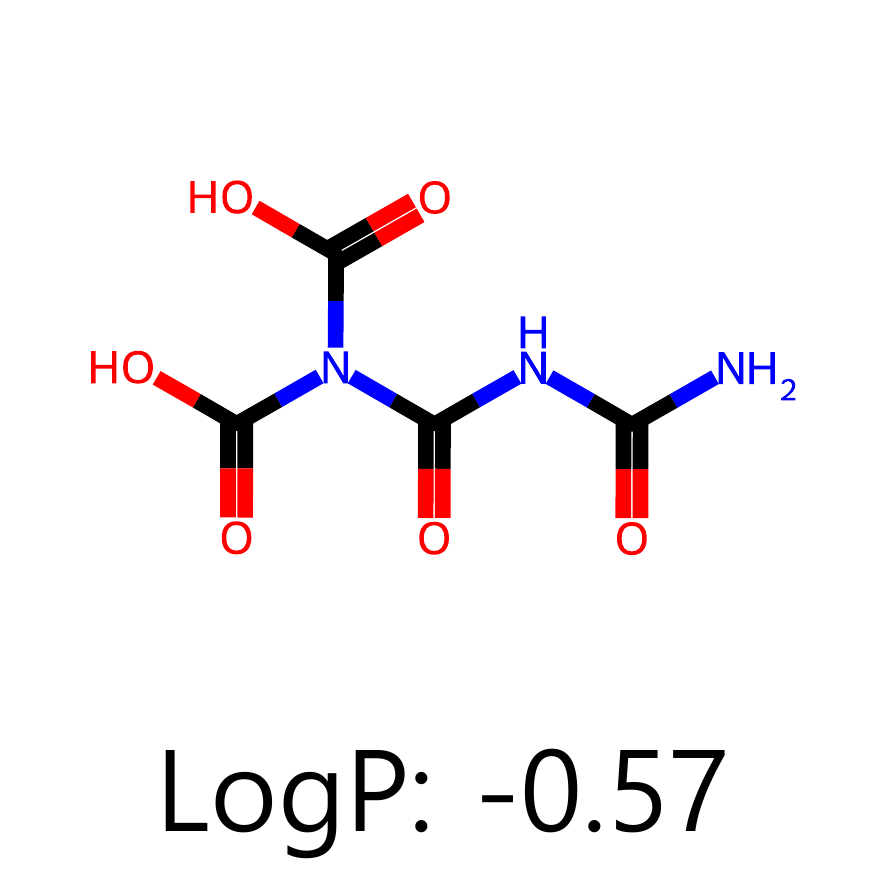}  & \includegraphics[height=0.7in,valign=c]{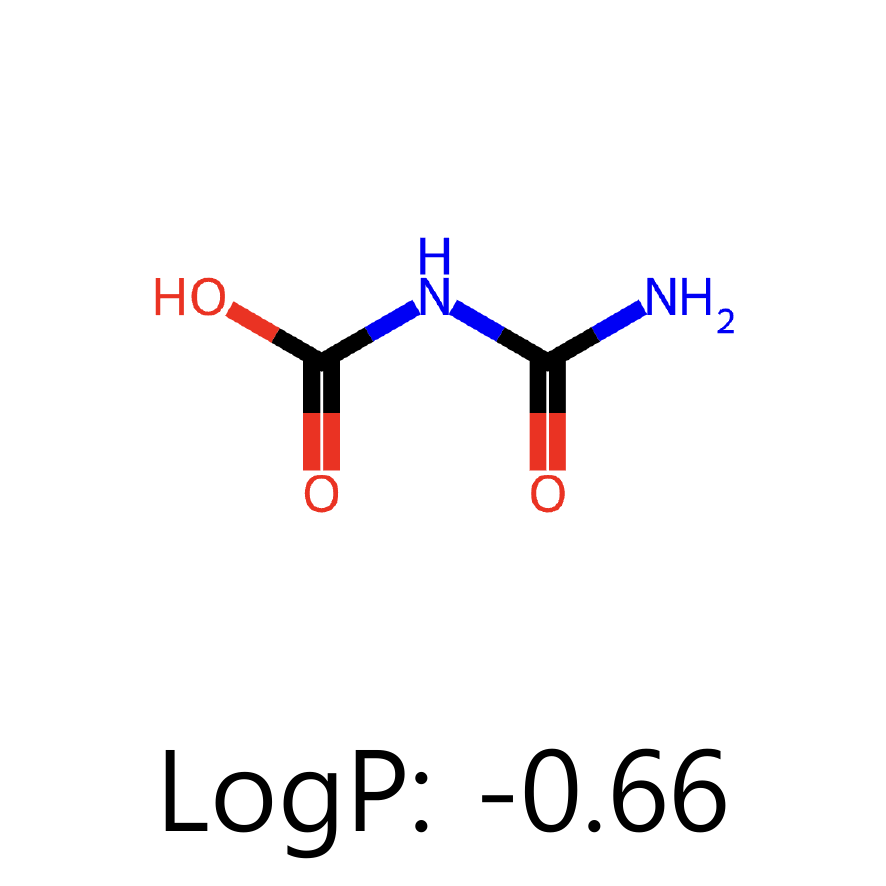}  & 
\includegraphics[height=0.7in,valign=c]{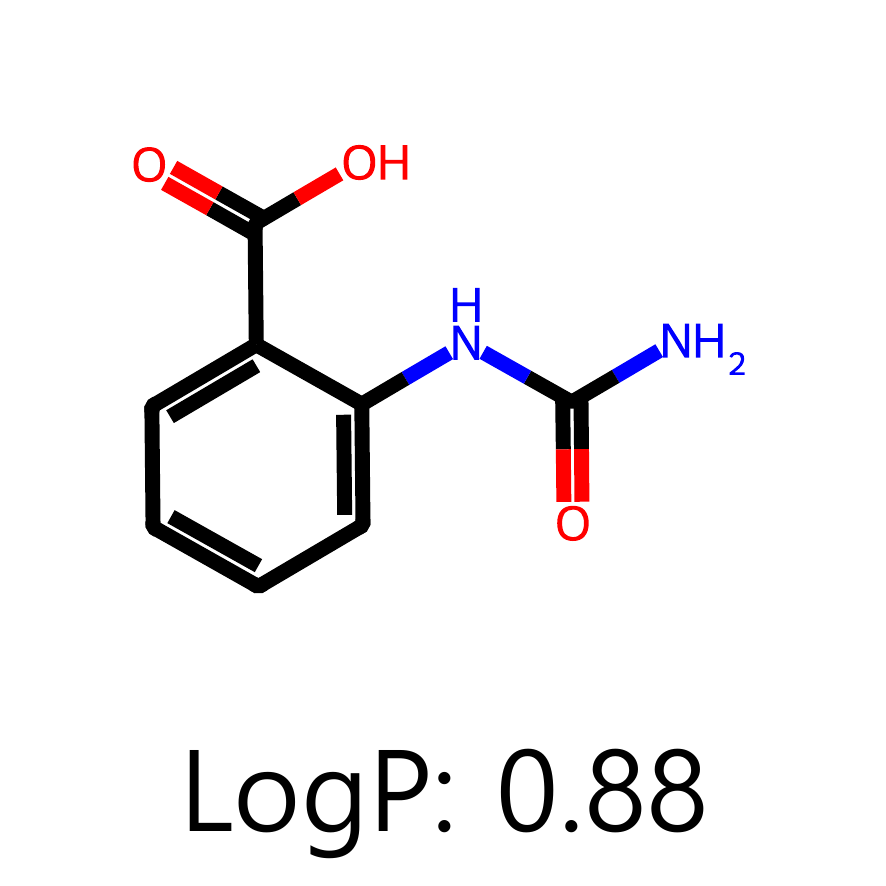} 
& 
\includegraphics[height=0.7in,valign=c]{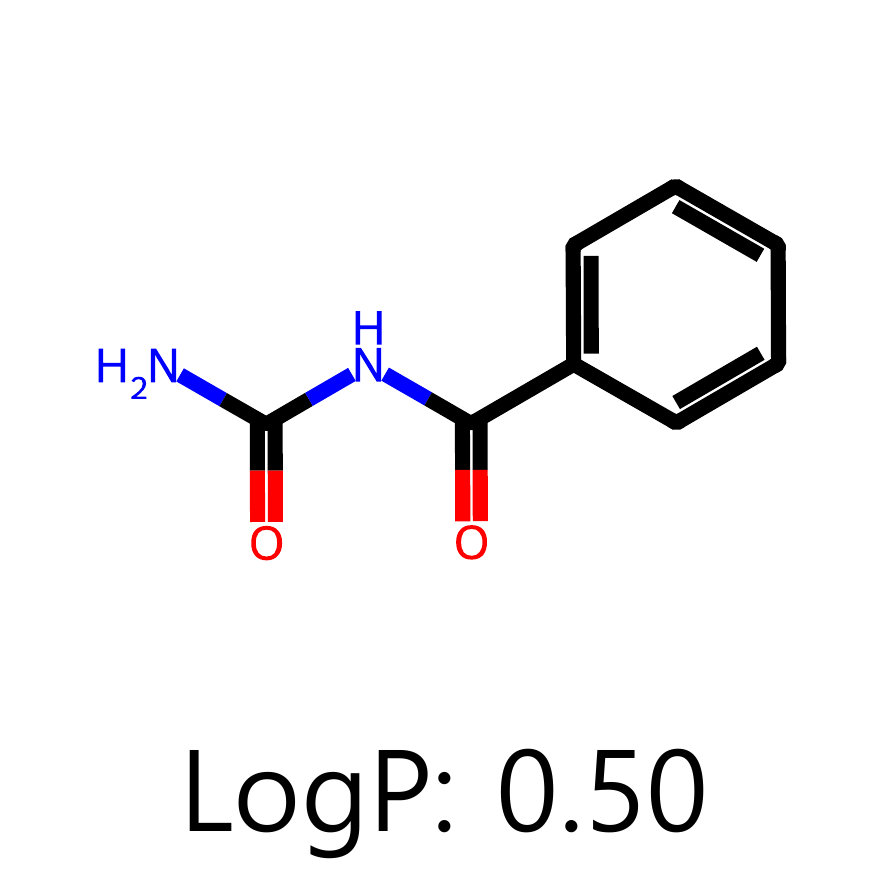}& 
\includegraphics[height=0.7in,valign=c]{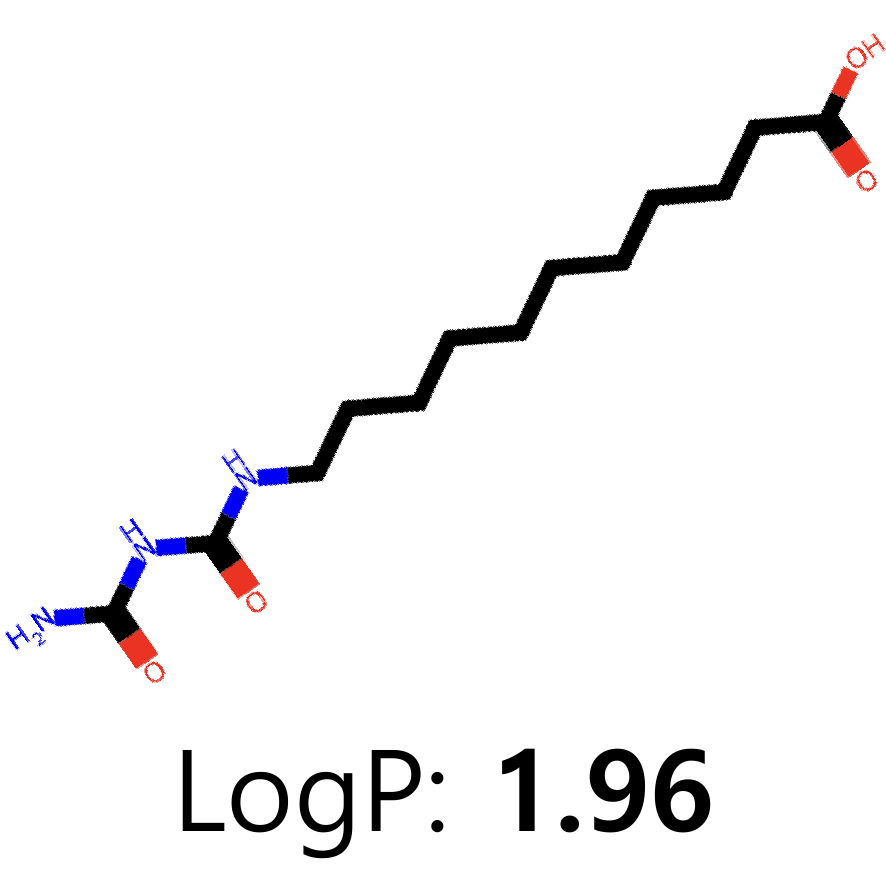} & 
\includegraphics[height=0.7in,valign=c]{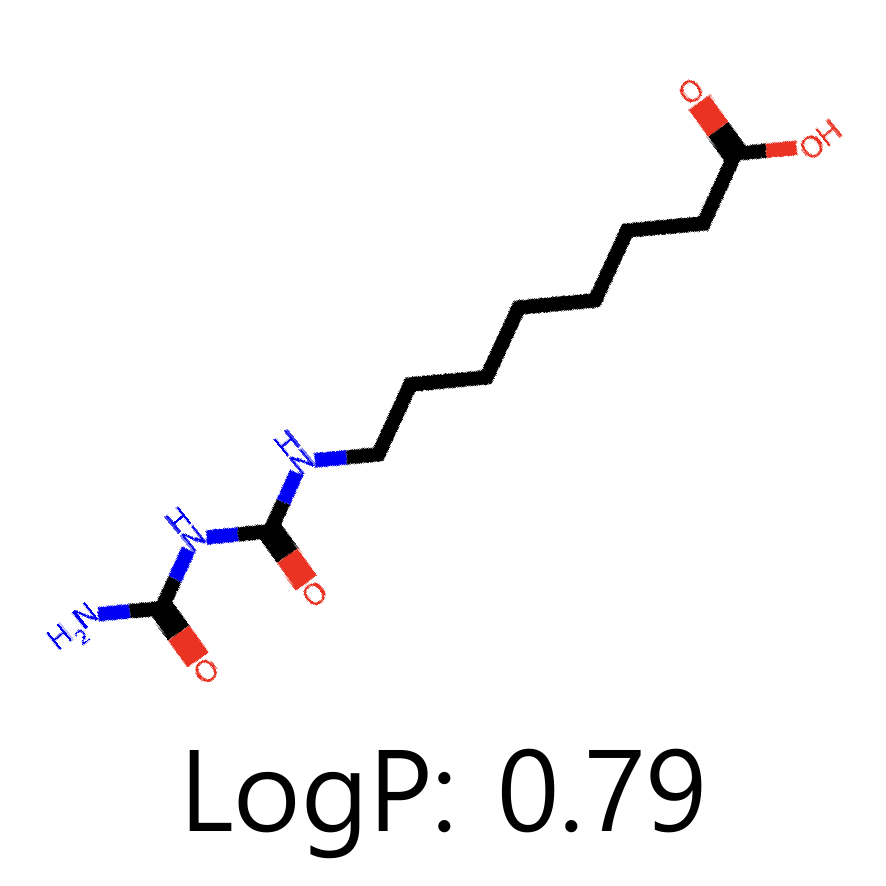} \\
  \bottomrule
\end{tabular}
\end{center}
\vspace{-0.15in}
\caption{
Qualitative results of molecule modification on ChEBI-20 \citep{edwards2021text2mol}. We visualize the generated molecules with respect to the prompt with an additional chemical condition, i.e., solubility in water. We report the LogP score below each visualization. Molecules with lower LogP values are more soluble in water. For the best-performing models of MolT5, BioT5, and CAMT5, we report the top-2 molecules that match the property description among 30 generated molecules based on temperature sampling with $\tau$=0.5. We set the best score in bold.
}

\label{fig:case_study}
\end{table*}

%% file: tables/molecule_modification.tex
\begin{table}[ht]
\centering
\resizebox{\linewidth}{!}{%
\begin{tabular}{lcc|cc}
\toprule
&\multicolumn{2}{c}{{``Soluble in water''}} & \multicolumn{2}{c}{{``Insoluble in water''}}\\\cmidrule{2-5}
\textbf{Model} & \textbf{MACCS} $\uparrow$ & \textbf{$\Delta$LogP} $\uparrow$ & \textbf{MACCS} $\uparrow$ & \textbf{$\Delta$LogP} $\downarrow$ \\
\midrule
$\text{MolT5}$   &  0.351 & 1.96 & 0.268 & -1.49 \\
$\text{BioT5}$  & 0.357 & 2.01 & 0.286 & -1.71  \\ \midrule
\rowcolor{tablegreen} \textbf{$\text{CAMT5}$} & \textbf{0.441} & \textbf{2.26} & \textbf{0.378} & \textbf{-1.98} \\
\bottomrule
\end{tabular}}
\vspace{-0.1in}
\caption{
Quantitative results of molecule modification on ChEBI-20 \citep{edwards2021text2mol}. We average the scores of top-2 molecules from the test set descriptions.
}
\label{tab:molecule_modification}
\end{table}

%% file: tables/importance_ablation.tex
\begin{figure*}[ht]
\begin{minipage}{0.68\textwidth}
\small
\begin{center}
\resizebox{\linewidth}{!}{%
\begin{tabular}{l|c|ccccc}
\toprule
\textbf{Model} & \textbf{Importance} &\textbf{Exact $\uparrow$} & \textbf{MACCS $\uparrow$} & \textbf{RDK $\uparrow$} & \textbf{Morgan $\uparrow$} & \textbf{Valid. $\uparrow$} \\ \midrule
$\text{MolT5}_\mathtt{base}^\dagger$ & - & 0.326 & 0.847 & 0.797 & 0.720 & 0.950 \\ 
$\text{BioT5}_\mathtt{base}^\dagger$ & - & 0.344 & 0.842 & 0.773 & 0.664 & \textbf{1.000} \\ \midrule
\rowcolor{tablegreen}\textbf{$\text{CAMT5}_\mathtt{base}$} & \textcolor{SJRed}\xmark & 0.397 & 0.868 & 0.819 & 0.725 & \textbf{1.000} \\
\rowcolor{tablegreen}\textbf{$\text{CAMT5}_\mathtt{base}$} & \textcolor{darkblue}\cmark & \textbf{0.422} & \textbf{0.882} & \textbf{0.834} & \textbf{0.742} & \textbf{1.000} \\
\bottomrule
\end{tabular}
}
\end{center}
\vspace{-0.15in}
\captionof{table}{Quantitative results on the CheBI-20 \citep{edwards2021text2mol} benchmark. $\dagger$ denotes that the model is trained with the same training configuration, e.g., training dataset, as ours. We mark Importance if the importance-based pre-training strategy (see Eq. (\ref{eq:confidence})) is applied. We bold the highest score.}
\label{tab:analysis}
\end{minipage}\hfill
\begin{minipage}{0.3\textwidth}\centering
\vspace{0.08in}
\includegraphics[width=\linewidth]{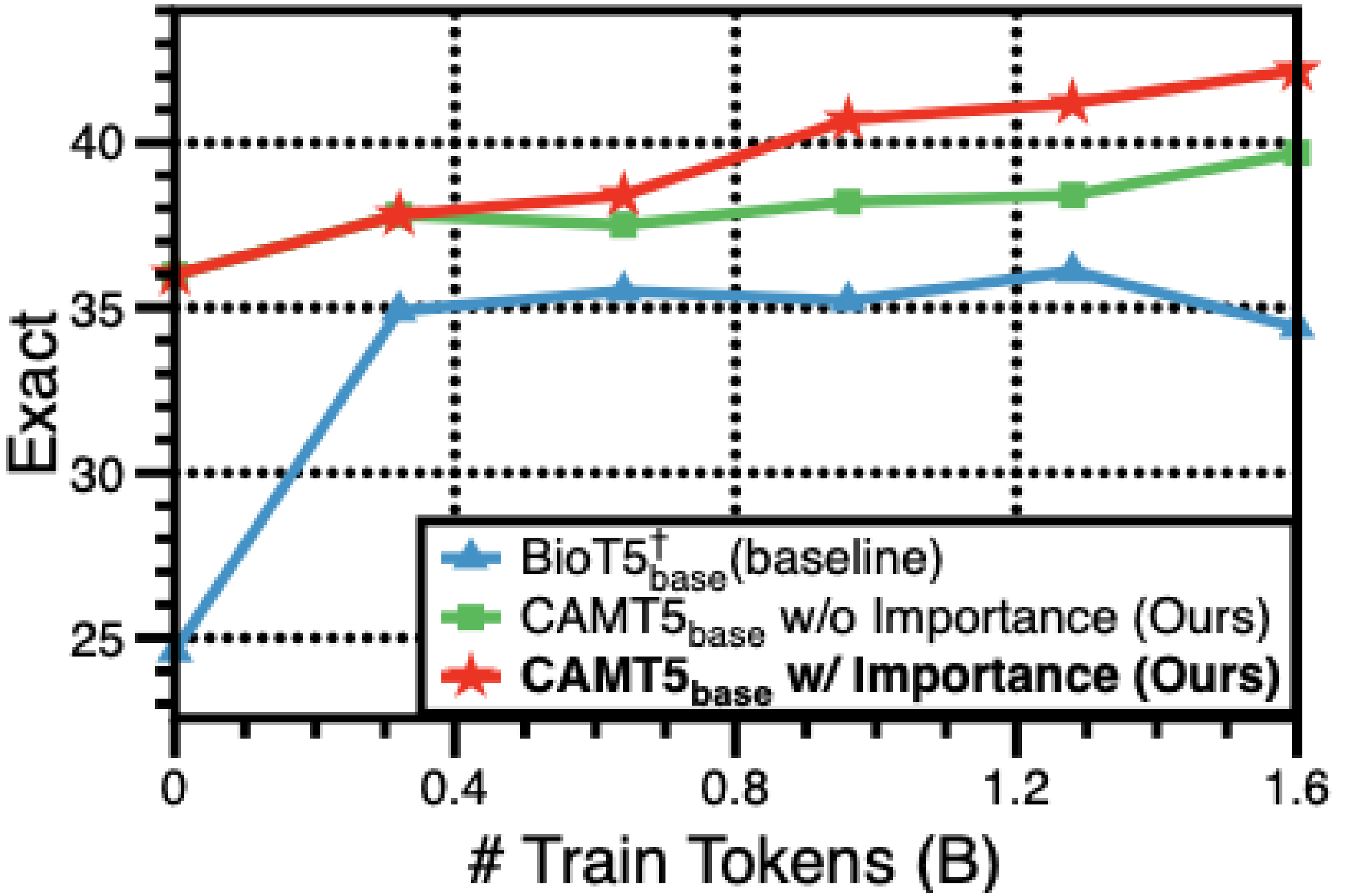} 
\vspace{-0.3in}

\caption{Performance varying the number of pre-training tokens.}
\label{fig:ablation_token_number}
    
\end{minipage}

    \end{figure*}

%% file: 5_conclusion.tex
\section{Conclusion}
\label{sec:conclusion}
We propose \Algname, a new text-to-molecule model with chemistry-specialized tokenization. Specifically, we adapt pre-trained language models by utilizing motif-level tokenization and importance-based training strategy to better understand the chemical structural context of molecules. In addition, we propose a confidence-based ensemble technique to further improve the quality of the generated molecules from \Algname, using other text-to-molecules. We hope that our work further accelerates future research on domain-specific adaptations of pre-trained language models.

%% file: 6_appendix.tex
\begin{center}
{\bf {\Large Appendix: Training Text-to-Molecule Models\\ with Context-Aware Tokenization}} 
\end{center}

\section{Context-aware tokenization details}
\label{sup: tokenization_details}
\input{figures/tokenization_example}

\input{figures/token_visualization}

In Figure~\ref{fig:tokenization_example}, we visualize the details of our context-aware tokenization scheme. For each motif-level token $M_i$, there may exist several $v\in V_i$ where $(u,v) \in \mathcal{E}$ for some $u\in V$, i.e., a single motif which is connected to several motifs in $\mathcal{T}$ (see the second token of \Algname in Figure~\ref{fig:concept} for an example). Therefore, we additionally store the order of fragmented bonds when traversing the motif tree. In the fragmented bonds in height 0, the marked number denotes the order of bonds that are connected to children in the linearized token sequence. In the fragmented bonds in height more than 0, the bond marked with zero is connected to its parent. Other bonds are connected to its children by their marked order, starting with 1. We also store the stereo information, e.g., tetrahedral or E-Z, in each token. We utilize this order when converting the sequence of tokens to a molecule. For a given sequence of tokens, we convert the sequence to a molecule by the exactly inverse consequences of the construction of the token sequences. If there exist unvisited fragmented bonds, we simply ignore them, i.e., we consider them to be connected to a hydrogen atom, not to other motif tokens. The number of motif tokens introduced in our \Algname is 24,735 in the ChEBI-20 and PCDes benchmarks. 

We visualize some of our motif tokens in {Figure~\ref{fig:token_visualization}}. We note that our choice of tokenization strategy is largely different from previous molecule representations. For example, t-SMILES \citep{wu2024t} constructs a full binary tree to construct a tree of motifs. However, such construction requires additional grammar representations, e.g., dummy nodes, to ensure the full binary trees. In contrast, our representation does not impose such restrictions, e.g., our motif token can have several children and thus does not require any grammar tokens.

\newpage
\section{Experimental details}
\label{sup:experimental_details}

{
\textbf{Details on token importance.} We simply use the number of atoms in each token as the importance value $\lambda(M_i)$: 
\begin{align}
\label{eq:tokenimportance}
\nonumber
    \lambda(M_i) = \mathtt{Softmax}(\log(A_i + 1)),
\end{align}
where $A_i$ denotes the number of atoms in each motif token. For special tokens, e.g., mask tokens, the atom count is set to 0.

\noindent\textbf{Details on pre-training.} For each text-to-molecule model, i.e., MolT5$^\dagger$, BioT5$^\dagger$, and CAMT5, we use a general text corpus (Colossal Clean Crawled Corpus \citep{raffel2020exploring}) and a molecule corpus (ZINC-15 \citep{sterling2015zinc}).
Specifically, each model is pre-trained on 1.6B of molecule-related tokens. Training is conducted with a batch size of 16 per GPU across 4 GPUs, with each batch containing an equal mix of text and molecule data. The training runs for 100k steps. We use AdamW with RMS scaling as the optimizer, and apply cosine annealing for the learning rate schedule. Gradients are clipped at 30.0. The base learning rate is 2e{-3}, and the number of warm-up steps is 1000. The maximum input length for pre-training is 512.
Except for our \Algname, the pre-training loss is the conventional masked language modeling loss from \citet{raffel2020exploring}.\footnote{Our importance-based pre-training strategy is not applicable for models with atom-level tokenization, since their tokens represent a single atom.} 

\noindent\textbf{Details on fine-tuning.} We fine-tune each model with description-molecule data pairs in the ChEBI-20 \citep{edwards2021text2mol} and the PCDes \citep{zeng2022deep} benchmarks. Additionally, we utilize 34k text-molecule pairs extracted from PubChem \citep{wang2009pubchem}, ensuring that no molecules overlap with those in the benchmarks.
Each model is trained based on the objective in Eq. (\ref{eq:training}) with the corresponding molecule token representations. We fine-tune the models in 50k steps with a batch size of 48, applying cosine annealing and gradient clipping at 30.0. We select the best model by varying the learning rate within [1e{-3}, 2e{-3}].
}

\noindent\textbf{Computing resources.} In our experiments, we use Intel(R) Xeon(R) Gold 6326Y CPU @ 2.90GHz. We use A6000 48GB GPUs for pre-training and a single NVIDIA GeForce RTX 3090 GPU for fine-tuning.

\newpage
\section{Dataset details}
\label{sup:dataset_details}
\input{figures/pubchem_visualization}

\input{figures/additional_visualization}

The ChEBI-20 dataset consists of 33,008 description-molecule pairs, split into 26,407/3,301/3,300 pairs for train/validation/test \citep{pei2023biot5}. The PCDes dataset contains more challenging 15,000 description-molecule pairs, split into 10,500/1,500/3,000 pairs for train/validation/test \citep{zeng2022deep}. Both are derived from qualified description-molecule pairs in the open-sourced PubChem database \citep{wang2009pubchem}, where each text description describes the corresponding molecule's structure and chemical properties. In Table~\ref{fig:our_dataset_visualization}, we provide some visualizations of our curated 34k text-molecule pairs, which are introduced in Section~\ref{subsec:experimental_setup}. In Table~\ref{fig:additional_visualization}, we visualize some description-molecule pairs of our main benchmark dataset, i.e., the ChEBI-20 \citep{edwards2021text2mol} and PCDes \citep{zeng2022deep} benchmarks.

\newpage

\section{Additional experiments and analyses}
\label{sup:additional_experiments}
\subsection{Qualitative results}

\input{figures/chebi_pcdes_visualization}

In Table~\ref{fig:chebi_pcdes_visualization}, we provide some visualizations of the generated molecules from each text-to-molecule model. From these visualizations, we observe that our \Algname effectively generates molecules that contain crucial motifs of the target molecules, e.g., {imidazole} in the second row, which further demonstrates the importance of our motif-level tokenization scheme in \Algname. 

\subsection{Results based on T5-large models}
\input{tables/large_models}
In Table~\ref{tab:large_results}, we report the results of the text-to-molecule models derived from the T5-large \citep{raffel2020exploring} backbone model. Our $\text{CAMT5}_\text{large}$ outperforms the previous state-of-the-art text-to-molecule models, improving the Exact score by 0.375 $\rightarrow$ 0.430.

\newpage
\subsection{Performance varying the size of molecules}
\input{tables/mol_size_performance}
In Table~\ref{tab:atomcount_results}, we analyze the generation performance on the ChEBI-20 benchmark grouped by the number of atoms in molecules. Our tokenization consistently outperforms other methods when working with both small and large molecules. This is likely due to the fact that our tokenization successfully incorporates both local and global molecular information.

\subsection{Performance on atom-level descriptions}
\input{tables/atom_level_description}

In Table~\ref{tab:atom_level_grouped}, we evaluate the generation performance on ChEBI-20 that include atom-level descriptions. CAMT5 consistently outperforms MolT5 and BioT5 across various atom-level descriptions such as `chlor', `fluoro', `phospho', and `sulf', demonstrating its robustness in handling atom-specific information.

\newpage

\subsection{Comparison with alternative tokenization strategies}
\label{sup:t-smiles}
\input{tables/frag_ablation}
In Table~\ref{tab:frag_ablation}, we compare our tokenization strategy with previously proposed motif-aware tokenizations i.e, t-SMILES~\citep{wu2024t} and BRICS~\citep{degen2008art}, following their official implementations. The result shows that our tokenization strategy achieves superior performance across all metrics.

\subsection{Analysis on linearization algorithms in tokenization}
\input{tables/search_ablation}
In Table~\ref{tab:search_ablation}, we compare the linearization algorithms used in our tokenization strategy (see Figure~\ref{fig:token_visualization}). We adopt depth-first search (DFS) as our traversal strategy, which is a common linearization algorithm in molecular serialization such as SMILES~\citep{weininger1988smiles}. To verify whether this choice is indeed effective, we compare DFS with breadth-first search (BFS), another popular traversal method. The result shows that DFS consistently outperforms BFS across all evaluation metrics. We think that the nature of DFS traversal, which sequentially explores long, connected motif chains, facilitates the model’s ability to capture the underlying structural patterns of molecules.

\subsection{Ablation on token importance}
\input{tables/importance_objective_ablation}
In Table~\ref{tab:importance_ablation}, we compare the results varying the definition of imporatnce in Eq. (\ref{eq:confidence}), i.e., frequency of atoms and frequency of motifs. Specifically, frequency-based importance is defined as the inverse frequency of atoms or motifs in the training data prioritizing rare components. We find that our original choice of importance, i.e., the number of atoms, is the most effective among the candidates.

\newpage
\subsection{Ensemble performance analysis}
\input{tables/ensemble_ablation_acc}
In Table~\ref{tab:ensemble_accuracy}, we compare the ensemble result based on different combinations of models. Intriguingly, the result shows that even the model with inferior performance, e.g., MolT5, is useful when it is ensembled with other models, i.e., the performance is improved by 0.462 $\rightarrow$ 0.472.

\subsection{Data-efficient molecular generation}
\input{tables/data_efficient_generation}

In Table~\ref{tab:data_effficient_generation}, we compare the results on data-efficient molecular generation, which an important application of text-to-molecule models. \Algname outperforms other text-to-molecule models, verifying the potential of \Algname to other molecule-related applications.

\newpage
\subsection{Statistical analysis}
\input{tables/statistical_analysis}

In Table~\ref{tab:statistical_analysis}, we report the mean and standard deviation values based on 3 independently trained text-to-molecule models. \Algname shows superior performance across the evaluation metrics, consistently achieving the higher average scores compared to the baselines. 

\subsection{Computational cost}
\input{tables/computational_cost}

In Table~\ref{tab:computational_cost}, we provide the computational cost, including training costs, memory requirements, and inference costs. By treating multiple atoms as a single motif token, \Algname significantly reduces the token sequence length, leading to improved training and inference efficiency.

\section{Impact statement}
This work will accelerate improvements in the field of text-to-molecule models, which will affect various applications such as drug discovery and material design. However, malicious usage of text-to-molecule models (including our models) may lead to a potential threat of generating harmful chemicals. We believe that safeguarding these models is an important future research direction, which is also widely studied in various domains \citep{openai2023gpt4}. We used AI assistants in coding and draft refinement, e.g., grammar check.

%% file: figures/tokenization_example.tex
\begin{figure*}[h]
\centering\small
\vspace{-0.15in}
\includegraphics[width=\linewidth]{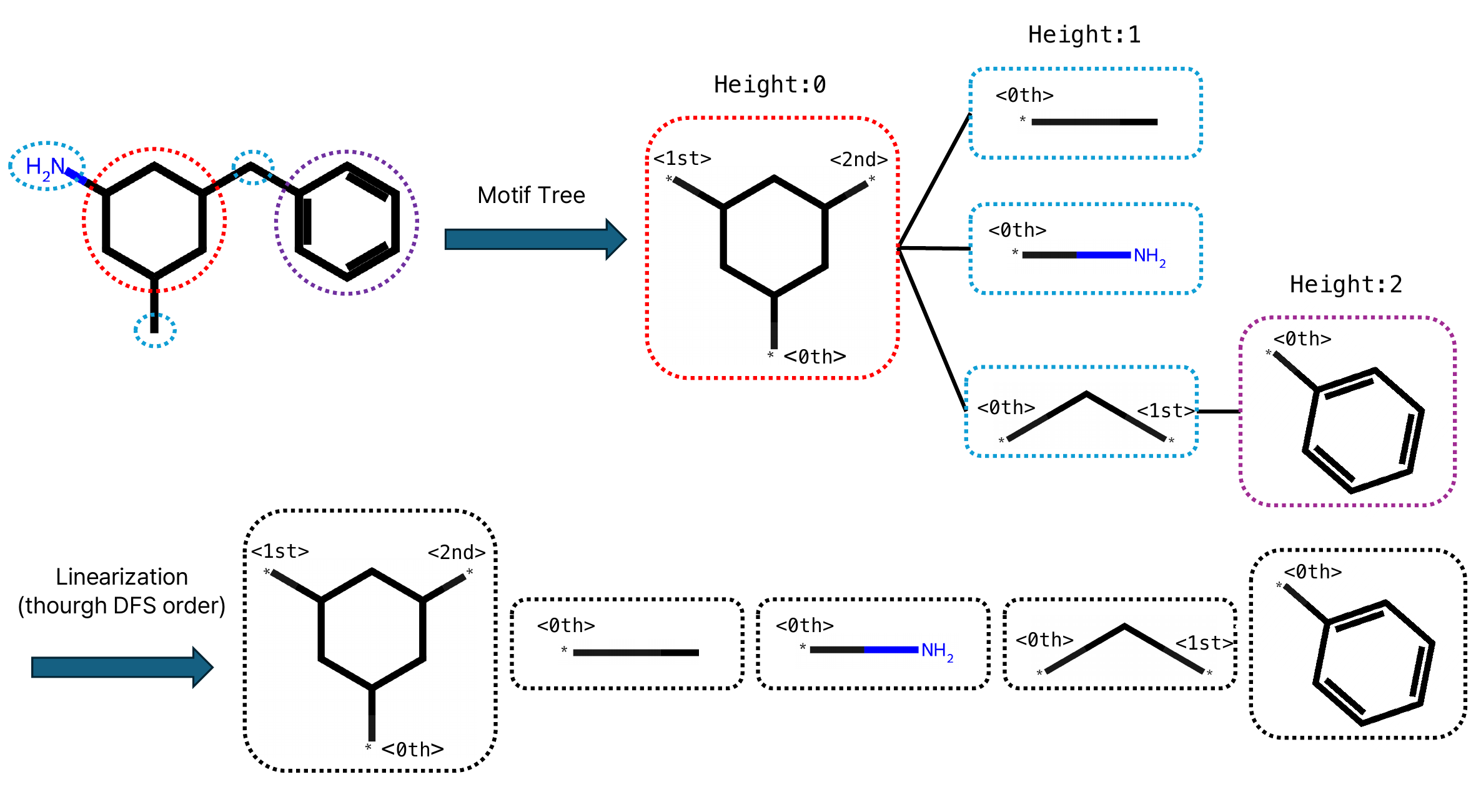}
\vspace{-0.3in}
\caption{
Details of our proposed tokenization scheme in \Algname: (1) atoms forming a ring structure, (2) atoms connected by a non-single bond, and (3) an atom not associated with (1) and (2) is considered as a single token.}
\label{fig:tokenization_example}
\end{figure*}

%% file: figures/token_visualization.tex
\vspace{-0.15in}
\begin{figure*}[h]
\centering\small
\includegraphics[width=6.00in]{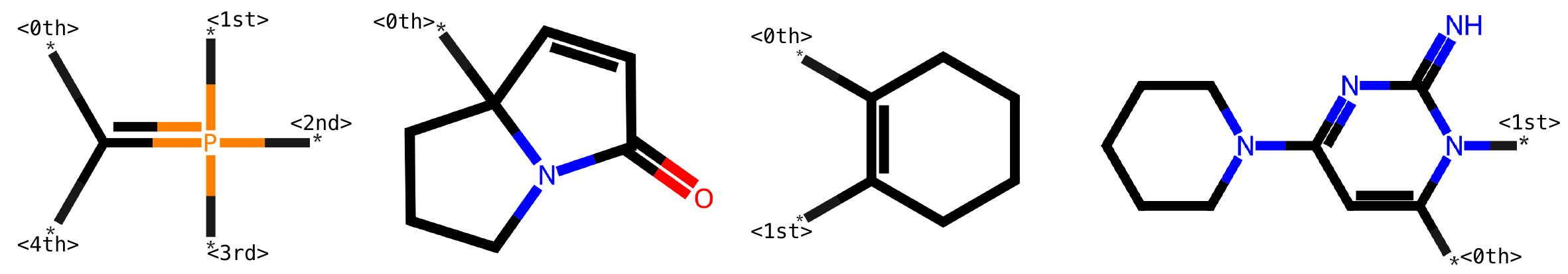}
\caption{
Visualizations of some context-aware motif tokens in \Algname.}
\label{fig:token_visualization}
\end{figure*}

%% file: figures/pubchem_visualization.tex
\begin{table*}[ht]
\begin{center}
\resizebox{1.0\textwidth}{!}{
\tiny
\begin{tabular}{c|c|c|c}
\toprule
  \multicolumn{4}{c}{PubChem} 
\\ \midrule
\includegraphics[height=0.7in,valign=c]{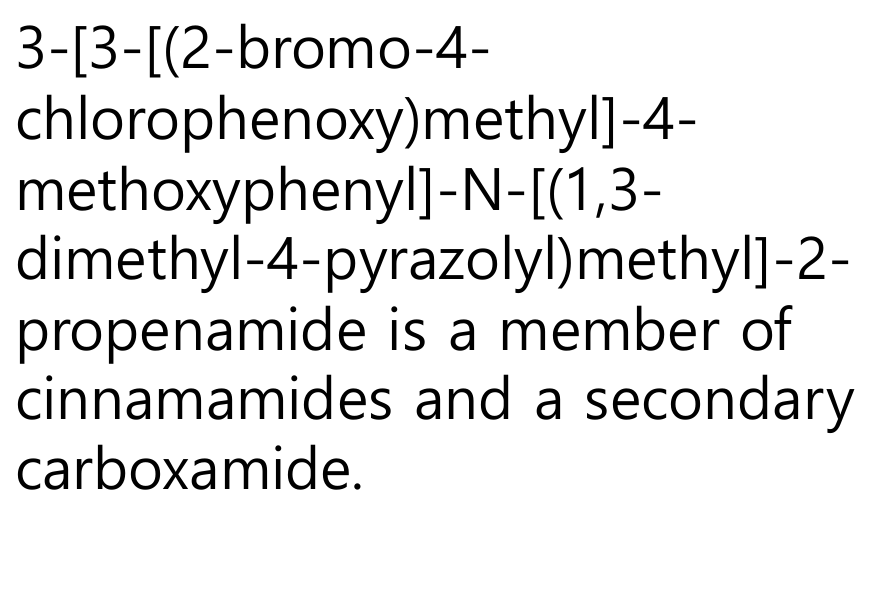}     &
\includegraphics[height=0.7in,valign=c]{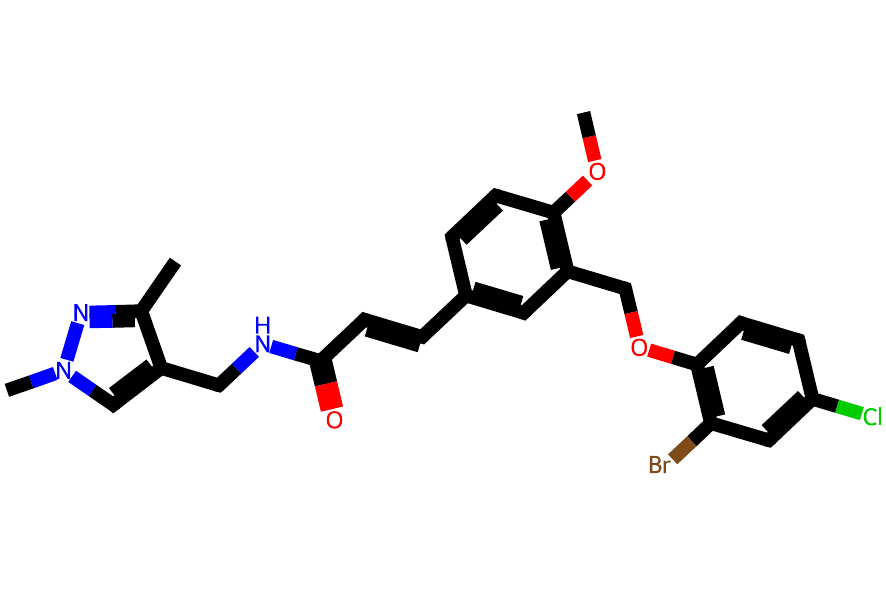} & 
\includegraphics[height=0.7in,valign=c]{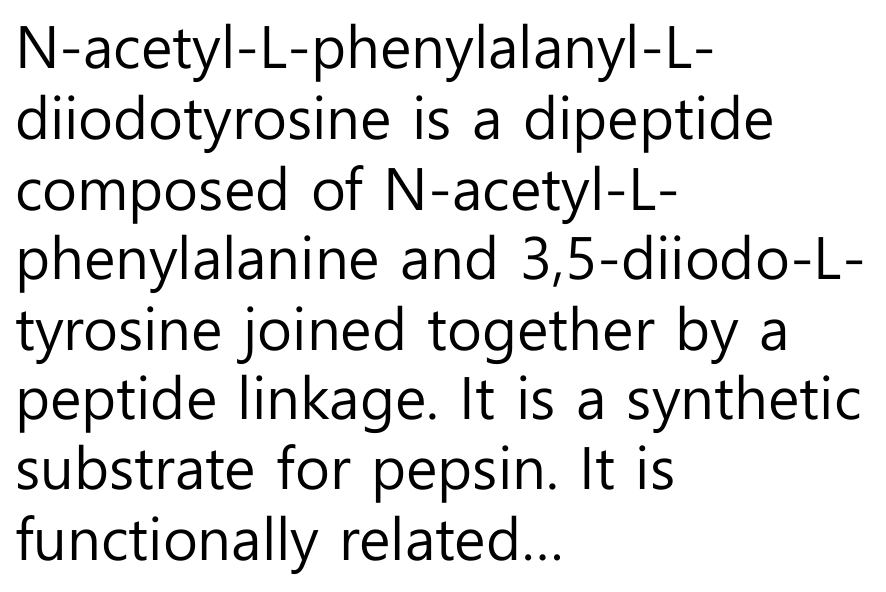}     &
\includegraphics[height=0.7in,valign=c]{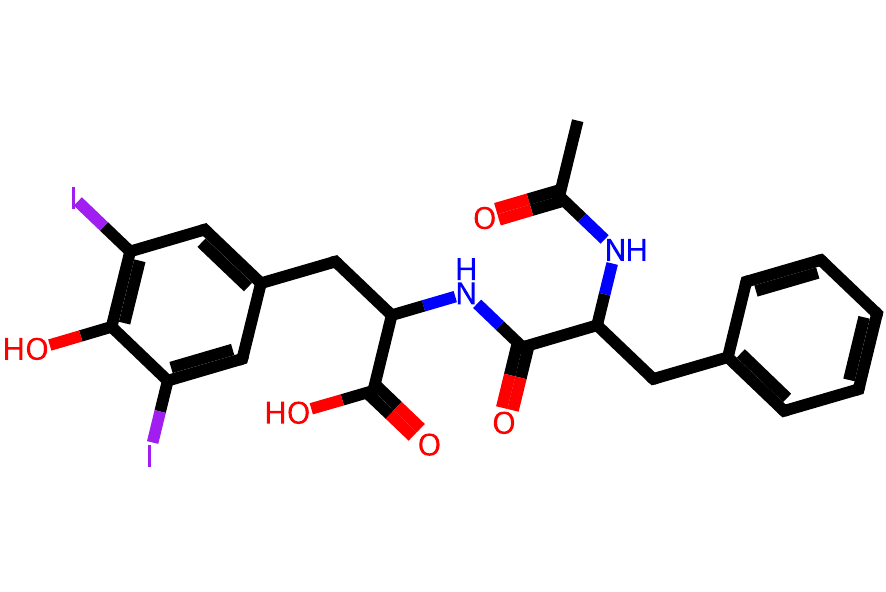} 

 \\\midrule
 \includegraphics[height=0.7in,valign=c]{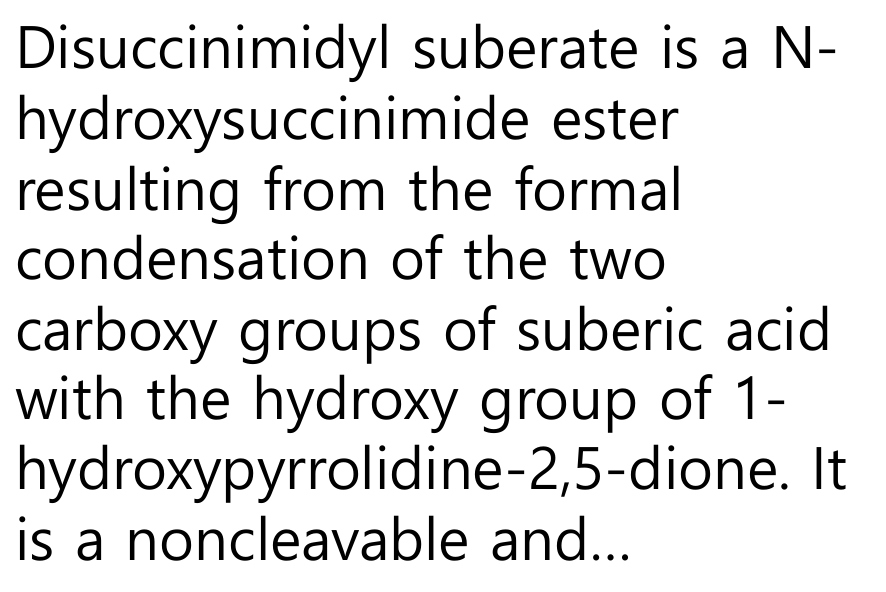}     &
\includegraphics[height=0.7in,valign=c]{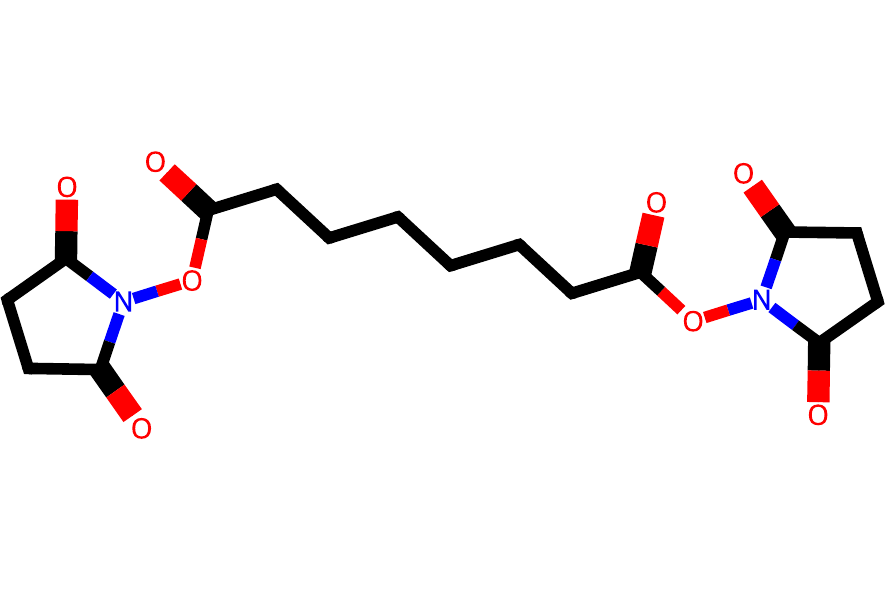} & 
\includegraphics[height=0.7in,valign=c]{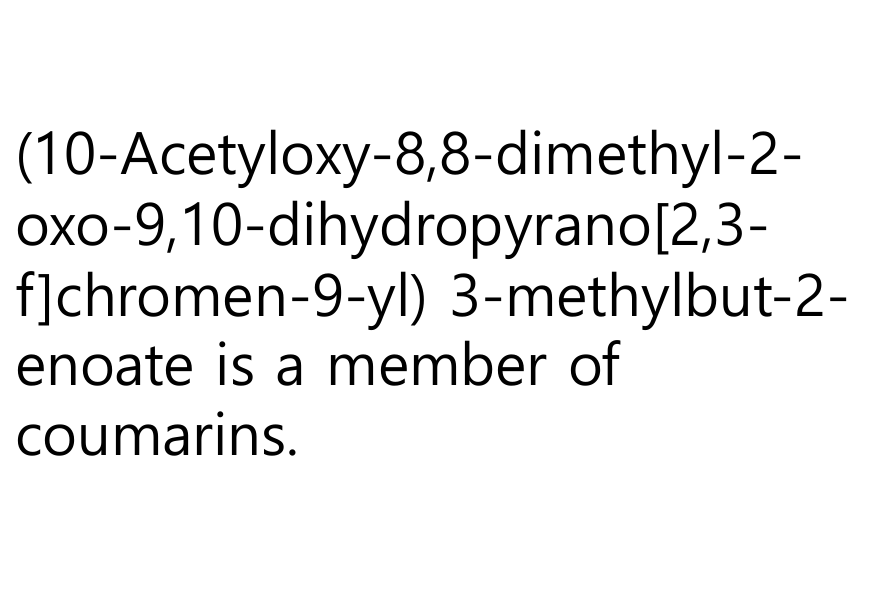}     &
\includegraphics[height=0.7in,valign=c]{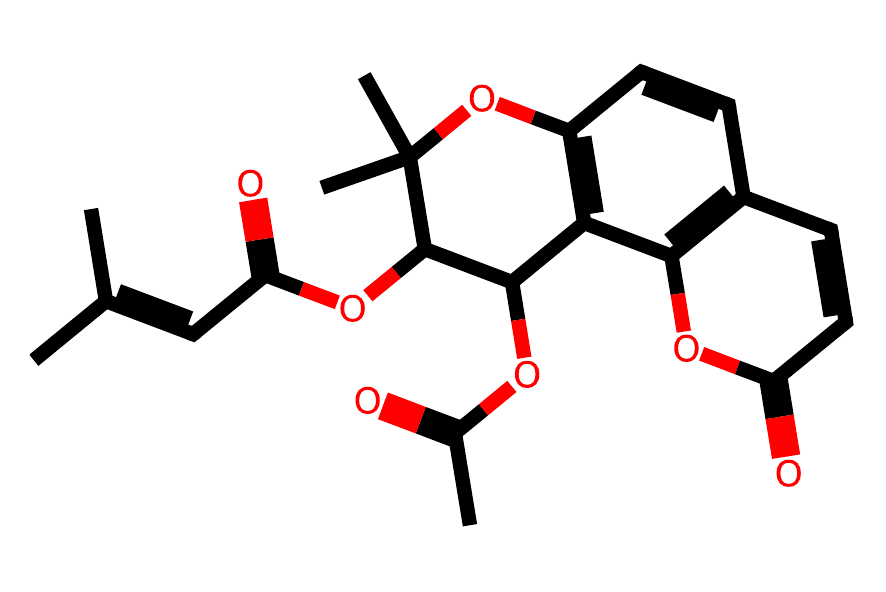} 

 \\
  \bottomrule
\end{tabular}}
\end{center}
\vspace{-0.15in}
\caption{
Visualizations of our description-molecule pairs collected from Pubchem database \citep{wang2009pubchem}.}
\label{fig:our_dataset_visualization}
\end{table*}

%% file: figures/additional_visualization.tex
\begin{table*}[ht]
\begin{center}
\resizebox{1.0\textwidth}{!}{
\tiny
\begin{tabular}{c|c|c|c}
\toprule
  \multicolumn{2}{c|}{ChEBI-20} & \multicolumn{2}{c}{PCDes} 
\\ \midrule
\includegraphics[height=0.7in,valign=c]{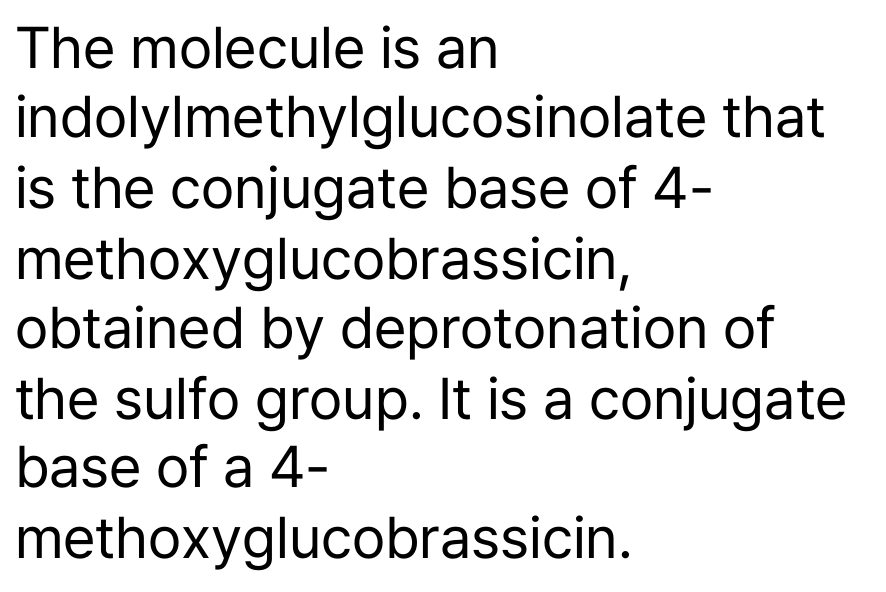}     &
\includegraphics[height=0.7in,valign=c]{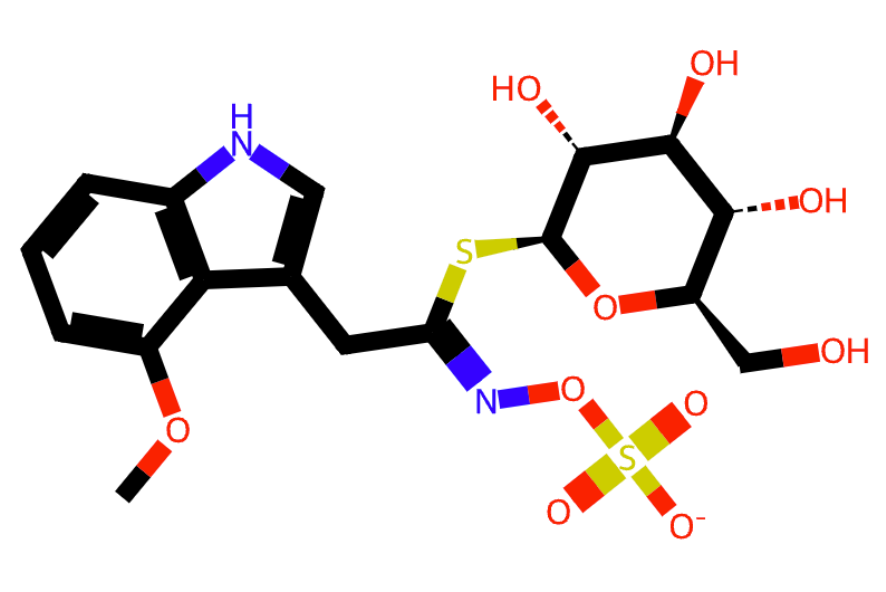} & 
\includegraphics[height=0.7in,valign=c]{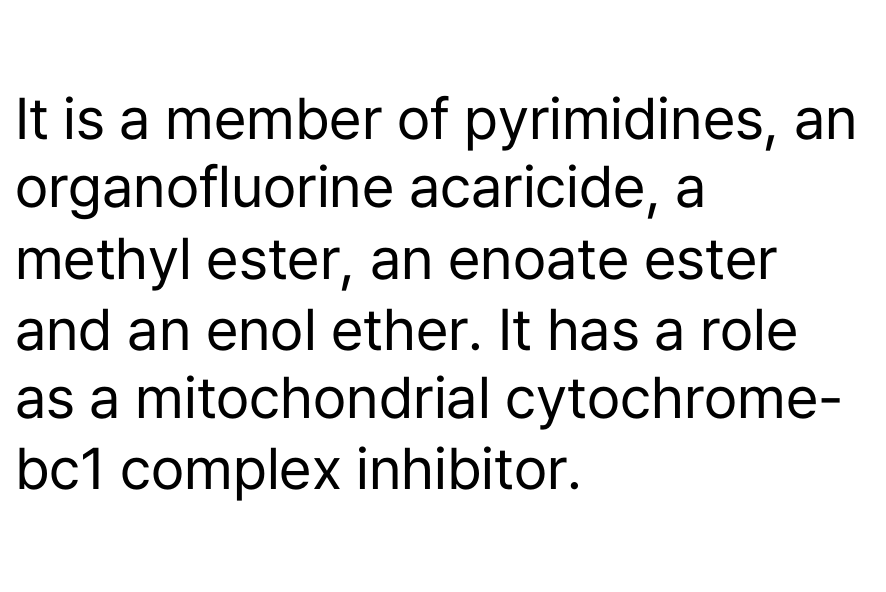}     &
\includegraphics[height=0.7in,valign=c]{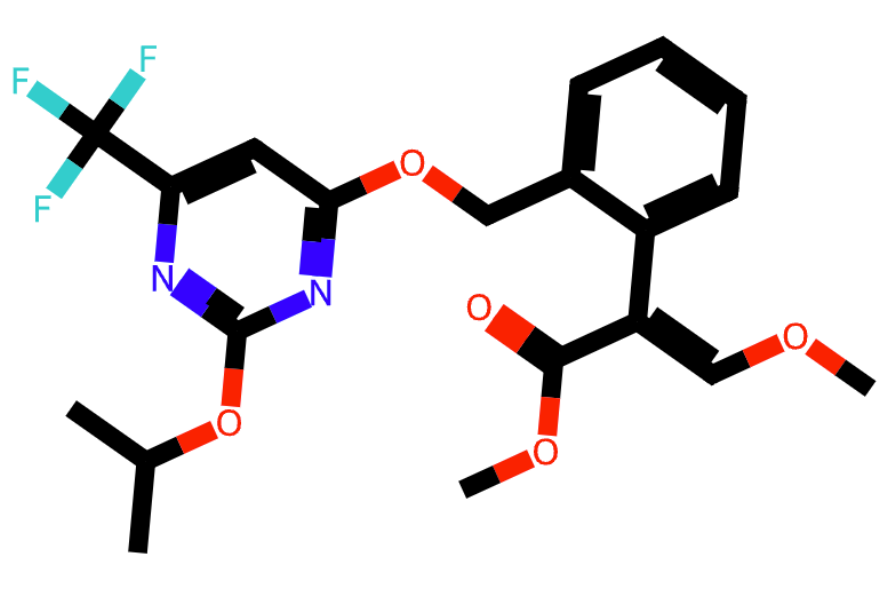} 

 \\\midrule
 \includegraphics[height=0.7in,valign=c]{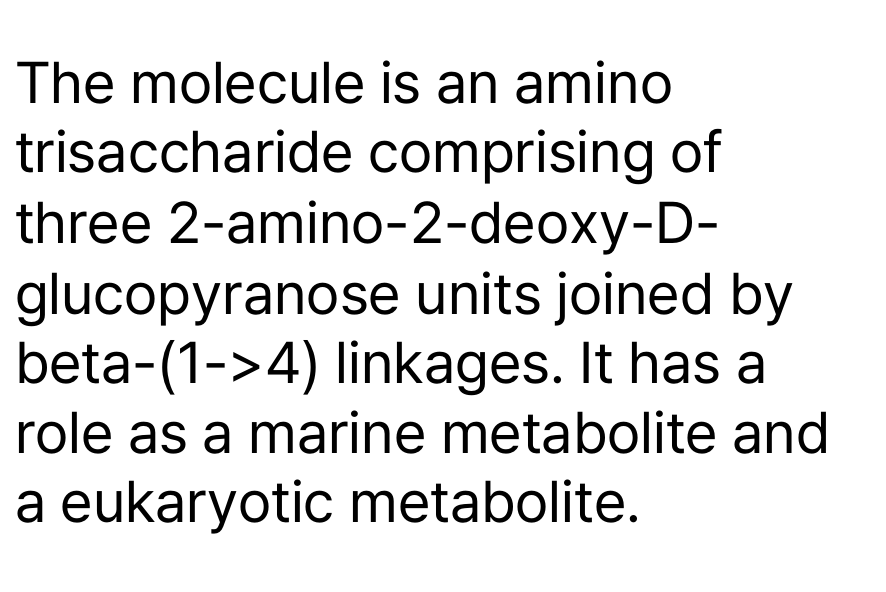}     &
\includegraphics[height=0.7in,valign=c]{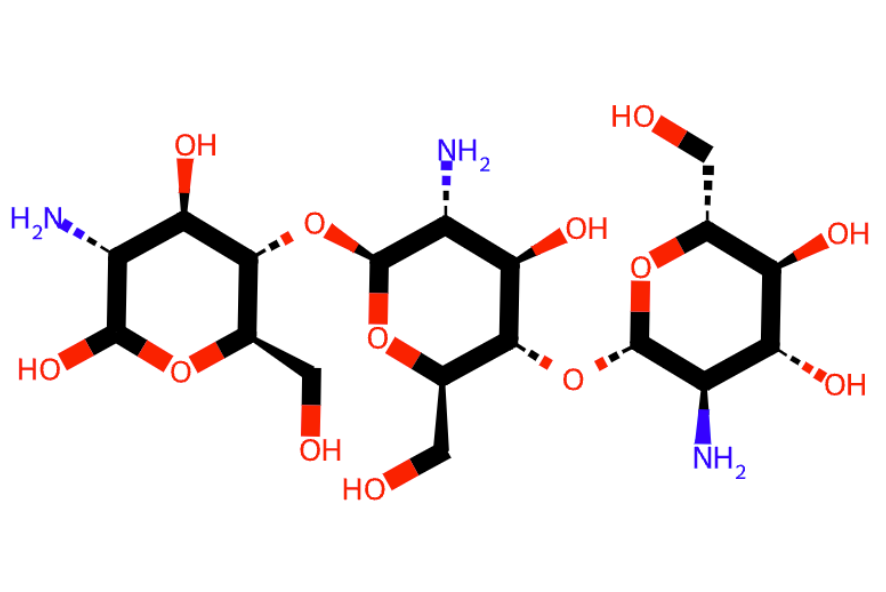} & 
\includegraphics[height=0.7in,valign=c]{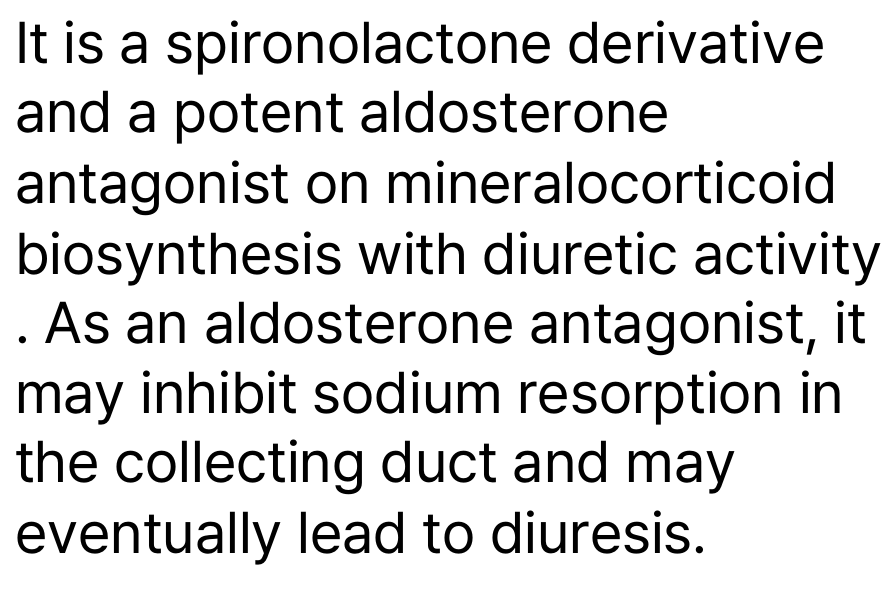}     &
\includegraphics[height=0.7in,valign=c]{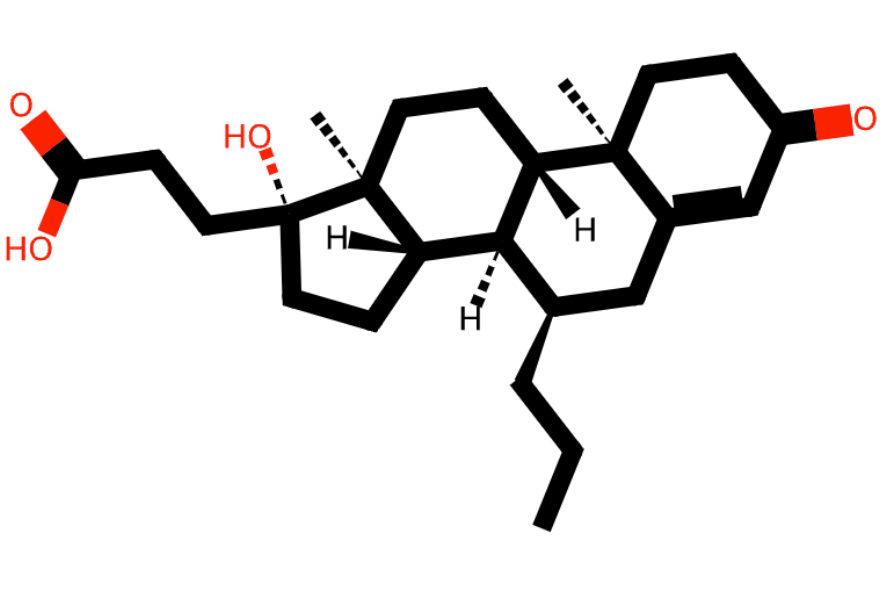} 

 \\\midrule
 \includegraphics[height=0.7in,valign=c]{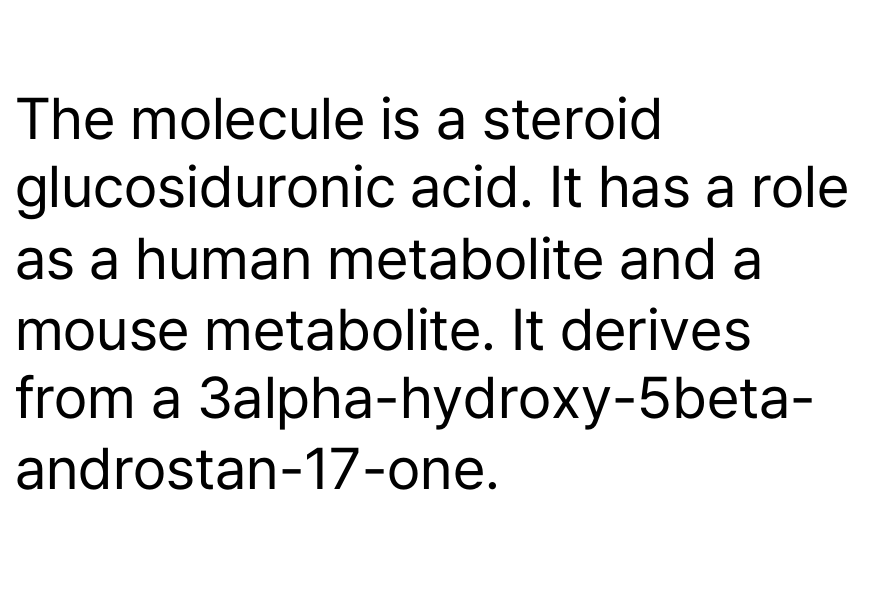}     &
\includegraphics[height=0.7in,valign=c]{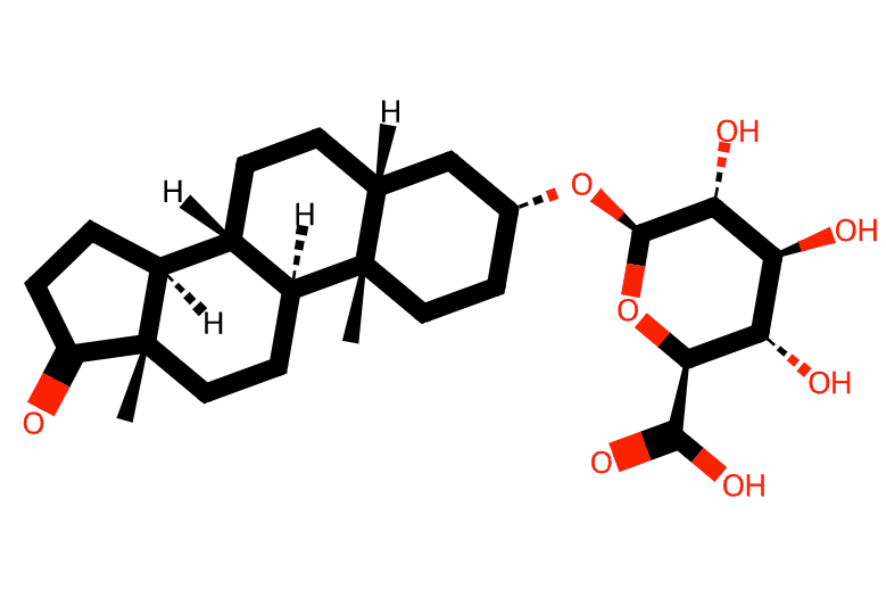} & 
\includegraphics[height=0.7in,valign=c]{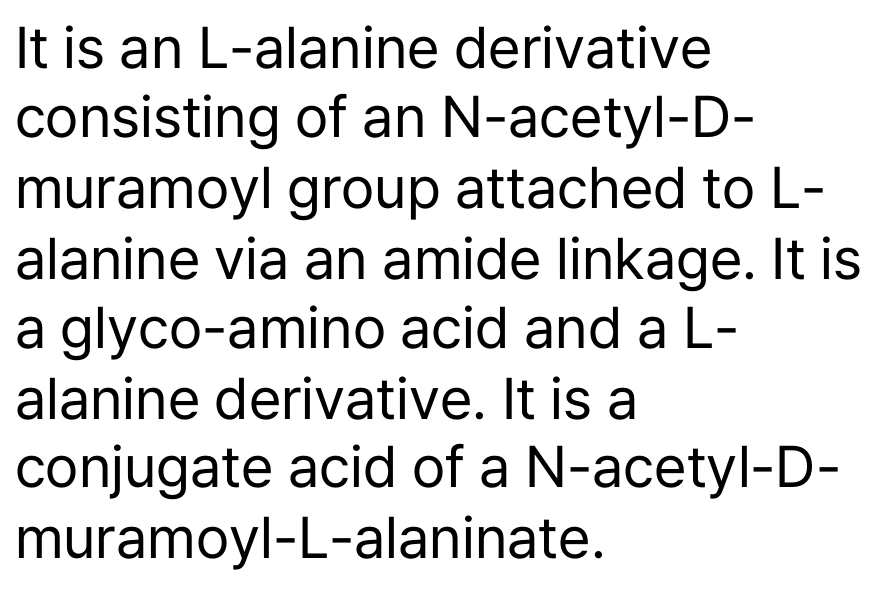}     &
\includegraphics[height=0.7in,valign=c]{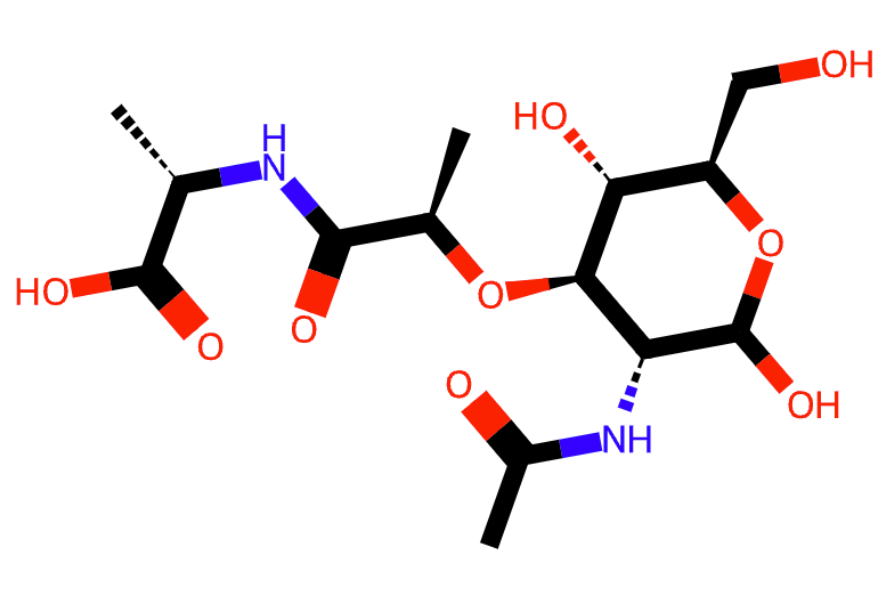} 

 \\
  \bottomrule
\end{tabular}}
\end{center}
\vspace{-0.15in}
\caption{
Visualizations of description-molecule pairs in the ChEBI-20 \citep{edwards2021text2mol} and PCDes \citep{zeng2022deep} datasets.}
\label{fig:additional_visualization}
\end{table*}

%% file: figures/chebi_pcdes_visualization.tex
\begin{table*}[ht]
\begin{center}
\small
\begin{tabular}{c|ccc|c}
\toprule
 Description &  $\text{MolT5}$ & 
$\text{BioT5}$  &
  \textbf{$\text{\Algname}$ (Ours)} & Target
\\ \midrule
\includegraphics[height=1.0in,valign=c]{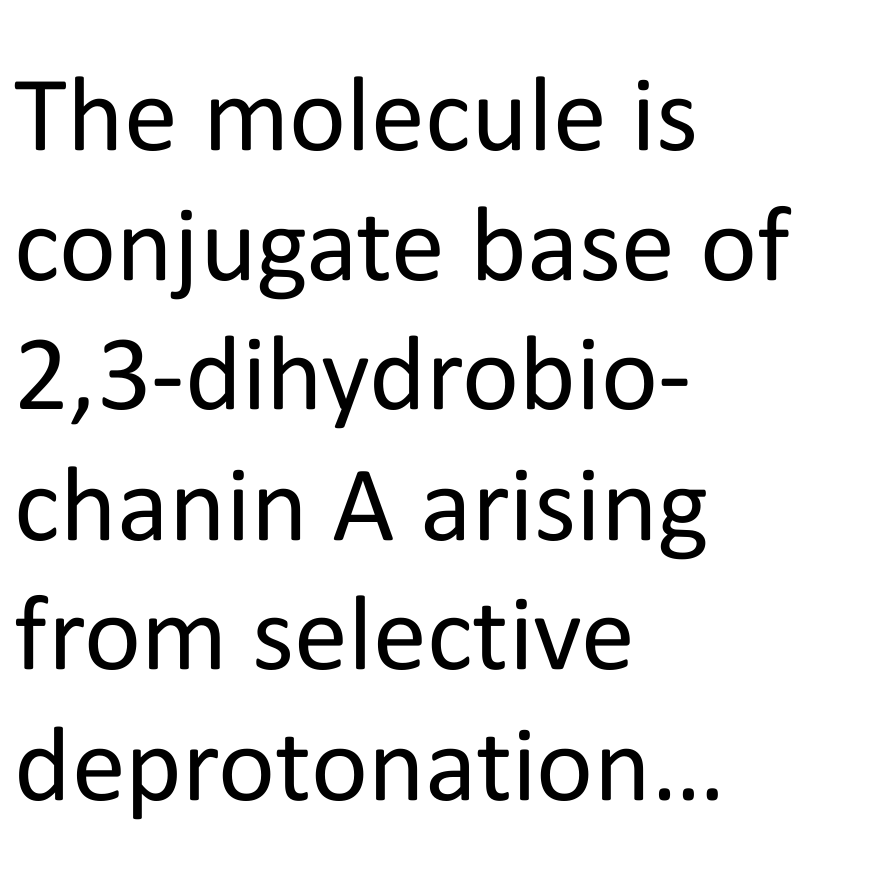}  &
\includegraphics[height=1.0in,valign=c]{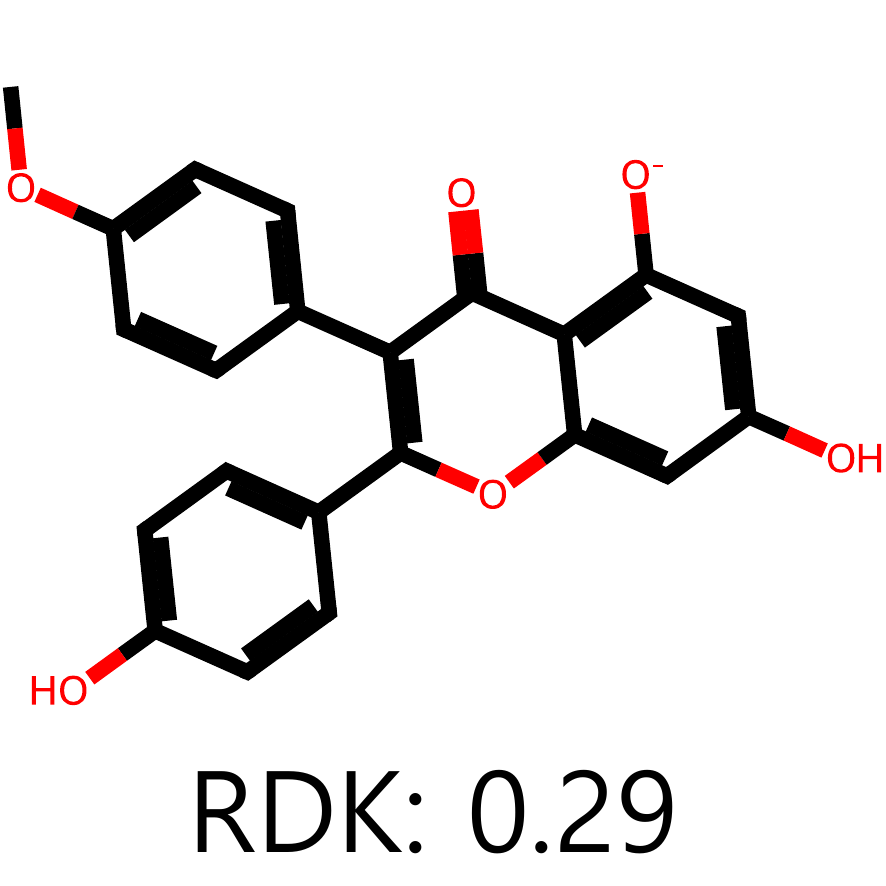} & 
\includegraphics[height=1.0in,valign=c]{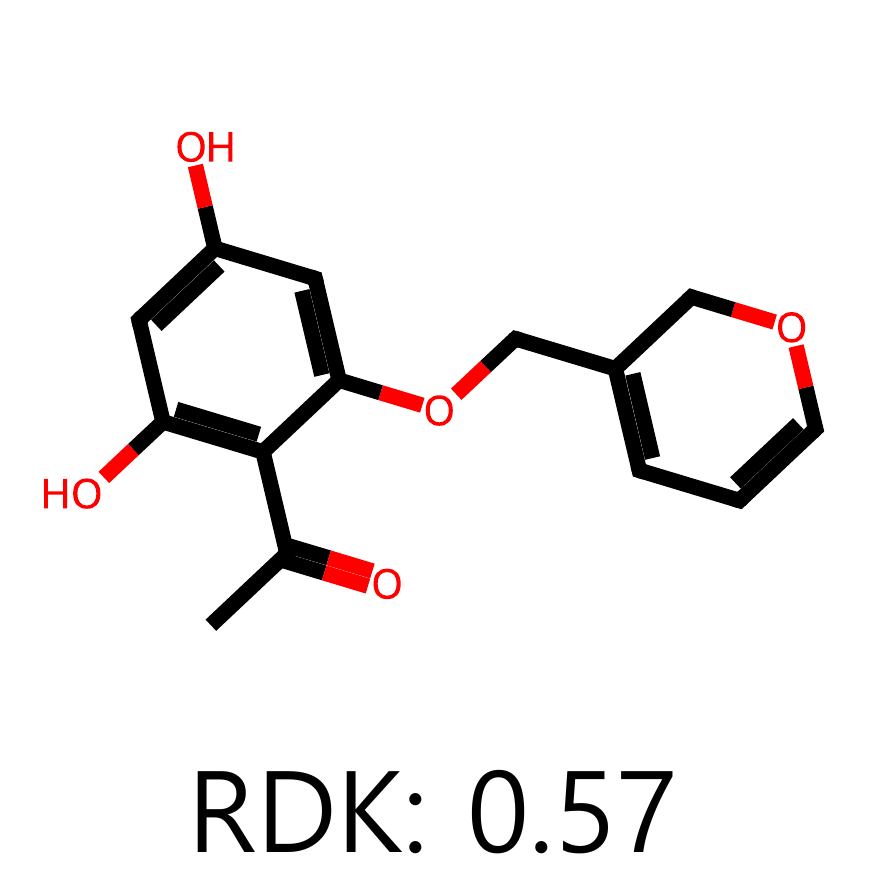}   &  
\includegraphics[height=1.0in,valign=c]{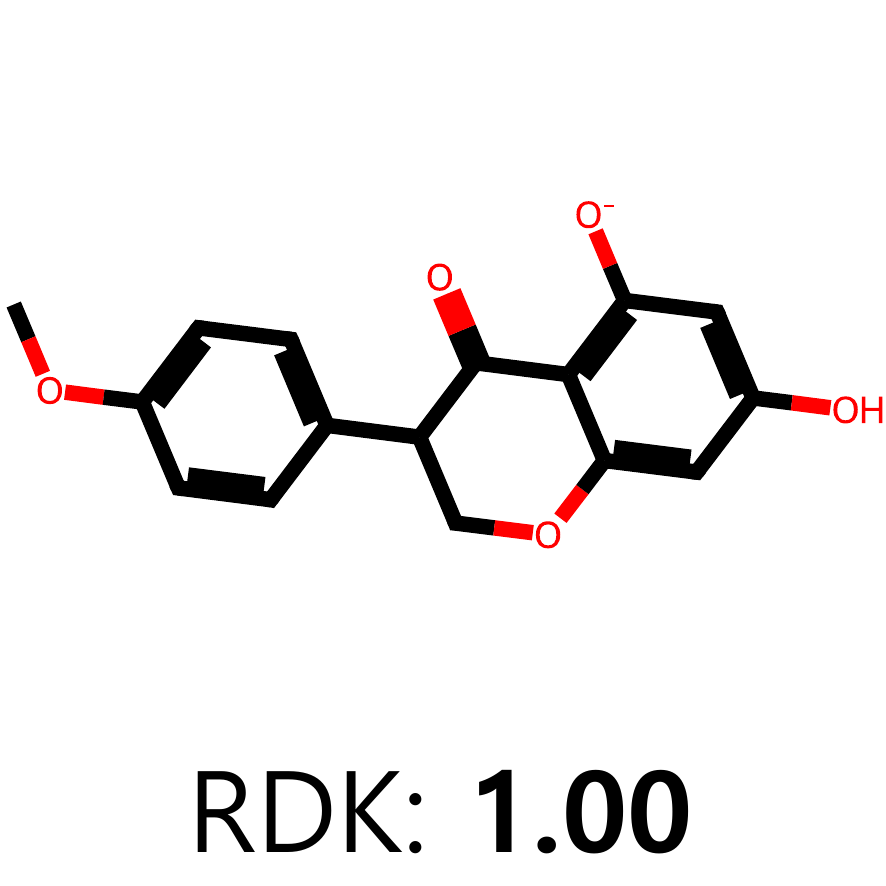} & 
\includegraphics[height=1.0in,valign=c]{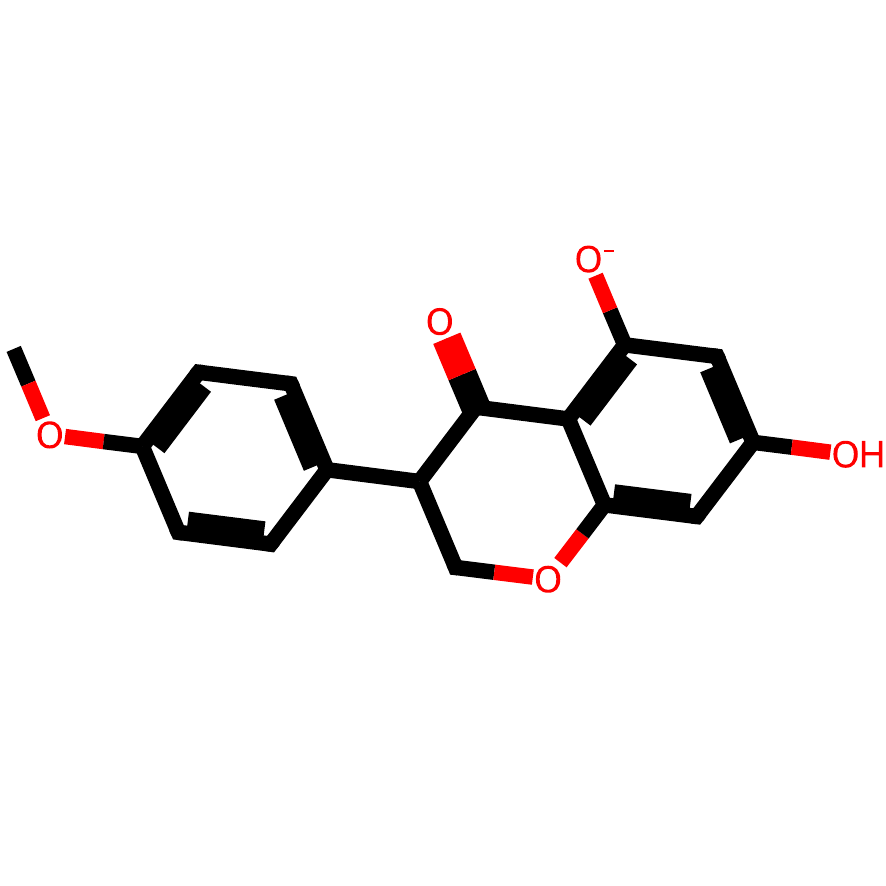}   \\\midrule
 \includegraphics[height=1.0in,valign=c]{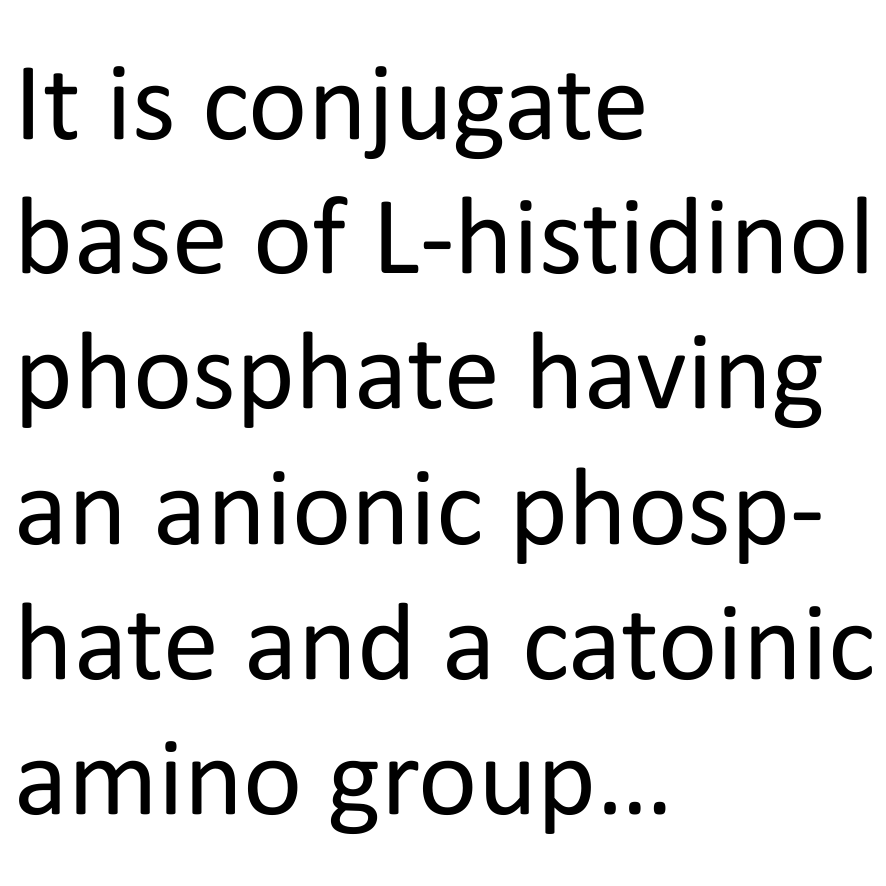} &   
\includegraphics[height=1.0in,valign=c]{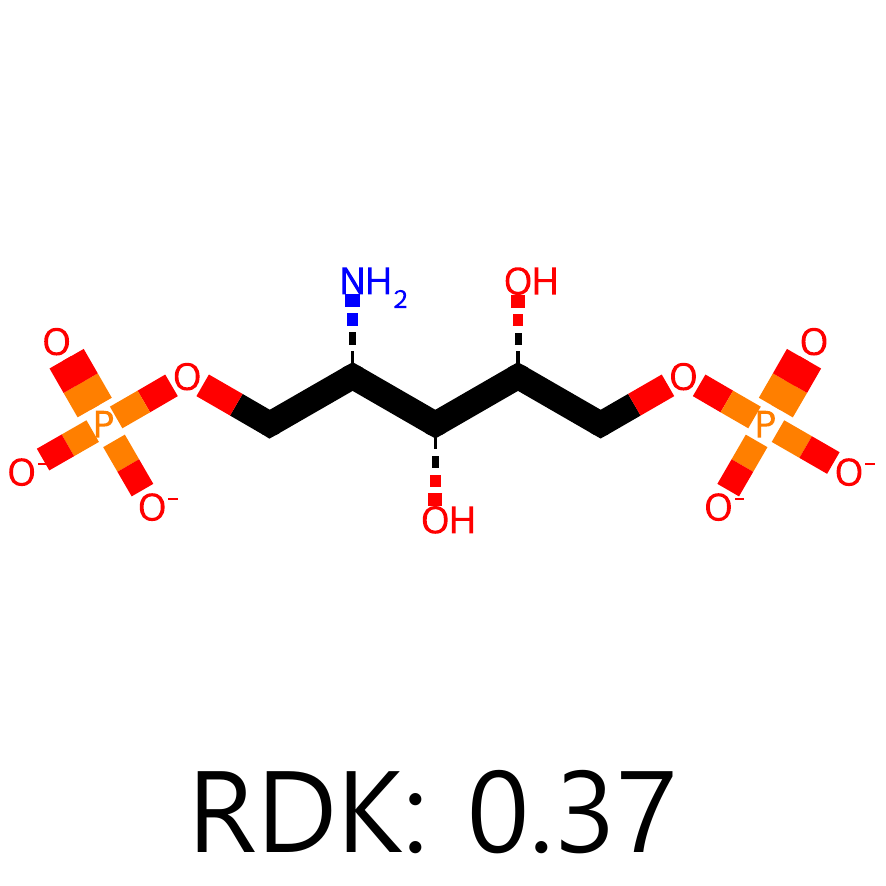} & 
\includegraphics[height=1.0in,valign=c]{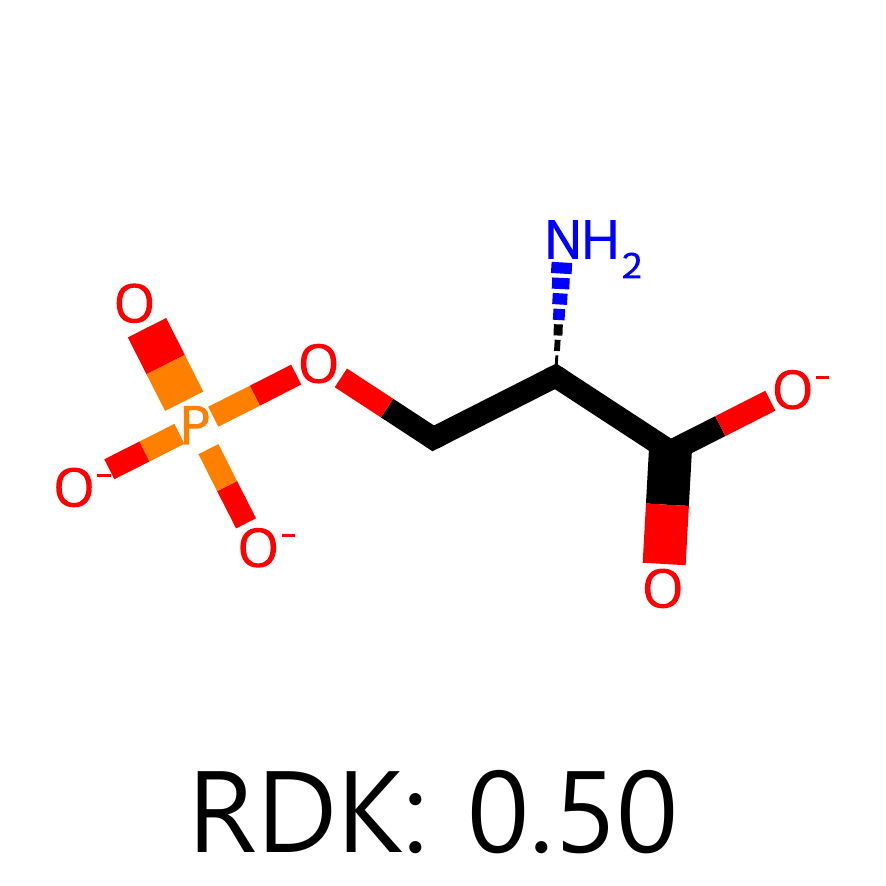} & 
\includegraphics[height=1.0in,valign=c]{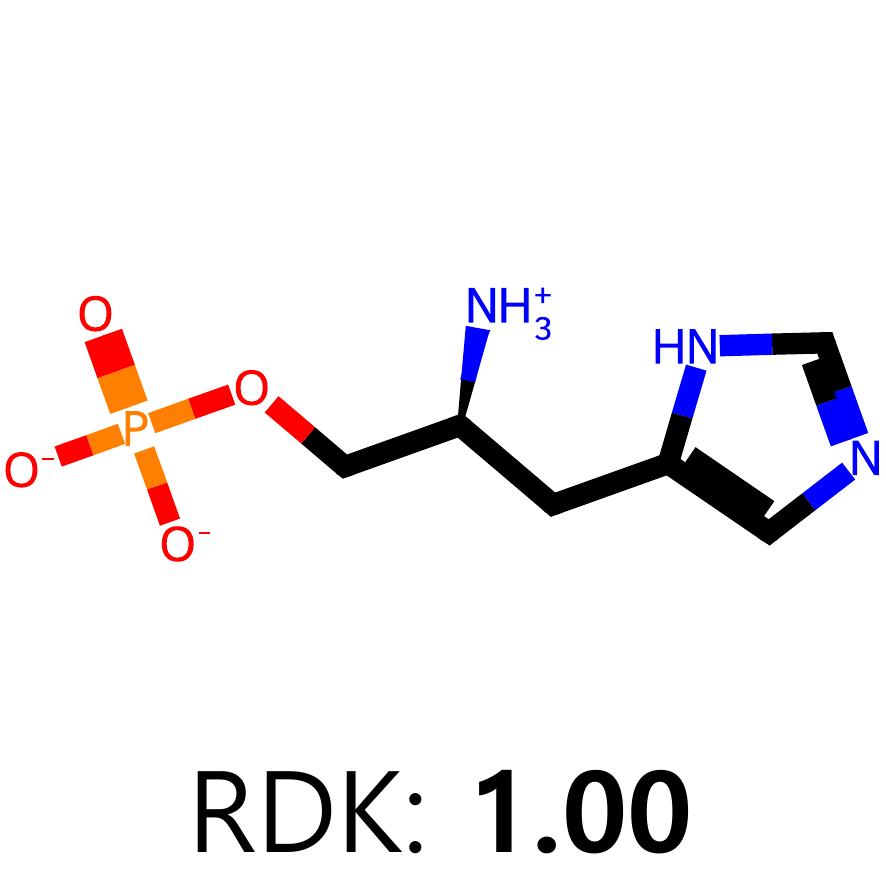} 
& 
\includegraphics[height=1.0in,valign=c]{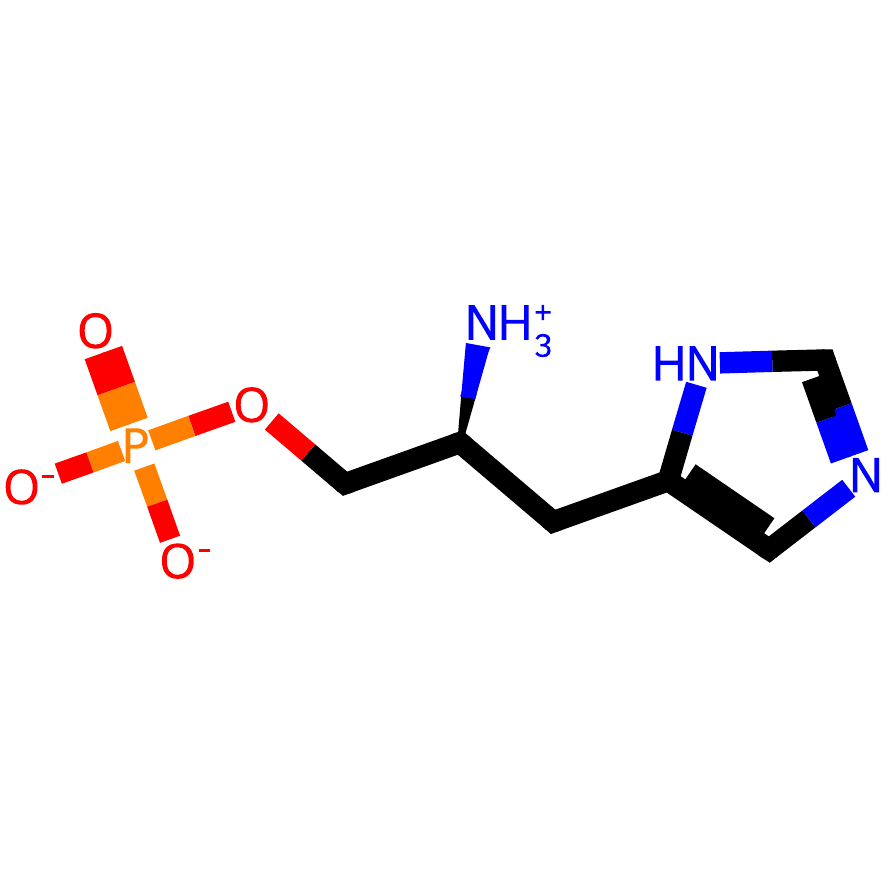} \\ 
  \bottomrule
\end{tabular}
\end{center}
\vspace{-0.15in}
\caption{
Qualitative results of the text-to-molecule generation task in the CheBI-20 \citep{edwards2021text2mol} benchmark (the first row) and PCDes \citep{zeng2022deep} benchmark (the second row). For the best-performing models of MolT5, BioT5, and CAMT5, we visualize the generated molecules with respect to the given description. We report the RDK score \cite{schneider2015get} between the generated and ground truth molecules below each visualization. We set the highest score in bold.
}
\label{fig:chebi_pcdes_visualization}
\end{table*}

%% file: tables/large_models.tex
\begin{table}[ht]
\centering

\begin{tabular}{l|ccccc}
\toprule
\textbf{Model} & \textbf{Exact} $\uparrow$ & \textbf{MACCS} $\uparrow$ & \textbf{RDK} $\uparrow$ & \textbf{Morgan} $\uparrow$ & \textbf{Validity} $\uparrow$ \\
\midrule
MolT5$_\text{large}^\dagger$ & 0.351 & 0.872 & 0.820 & 0.746 & 0.963 \\
BioT5$_\text{large}^\dagger$ & 0.375 & 0.855 & 0.790 & 0.688 & \textbf{1.000} \\ \midrule
\textbf{CAMT5$_\text{large}$ (Ours)} & \textbf{0.430} & \textbf{0.885} & \textbf{0.840} & \textbf{0.749} & \textbf{1.000} \\
\bottomrule
\end{tabular}
\vspace{-0.1in}
\caption{
Quantitative results on the ChEBI-20 \citep{edwards2021text2mol} benchmark. $\dagger$ denotes that the model is trained with the same training configuration, e.g., training dataset, as ours. We highlight the best score in bold. 
}
\label{tab:large_results}
\end{table}

%% file: tables/mol_size_performance.tex
\begin{table}[ht]
\centering
\begin{tabular}{l|l|ccccc}
\toprule
\textbf{\# Atoms} & \textbf{Model} & \textbf{Exact} $\uparrow$ & \textbf{MACCS} $\uparrow$ & \textbf{RDK} $\uparrow$ & \textbf{Morgan} $\uparrow$ & \textbf{Validity} $\uparrow$ \\
\midrule

\multirow{3}{*}{$(0, 30]$}
& MolT5$_\text{base}^\dagger$           & 0.365 & 0.848 & 0.789 & 0.708 & 0.964 \\
& BioT5 $_\text{base}^\dagger$             & 0.346 & 0.830 & 0.760 & 0.650 & \textbf{1.000} \\
& \textbf{CAMT5$_\text{base}$}  & \textbf{0.439} & \textbf{0.861} & \textbf{0.808} & \textbf{0.717} & \textbf{1.000} \\
\midrule

\multirow{3}{*}{$(30, 70]$}
& MolT5$_\text{base}^\dagger$           & 0.278 & 0.873 & 0.840 & 0.763 & 0.939 \\
& BioT5$_\text{base}^\dagger$           & 0.341 & 0.884 & 0.829 & 0.721 & \textbf{1.000} \\
& \textbf{CAMT5$_\text{base}$}  & \textbf{0.397} & \textbf{0.913} & \textbf{0.870} & \textbf{0.775} & \textbf{1.000} \\
\midrule

\multirow{3}{*}{$(70, \infty)$}
& MolT5$_\text{base}^\dagger$           & 0.194 & 0.826 & 0.811 & 0.749 & 0.854 \\
& BioT5$_\text{base}^\dagger$           & 0.291 & 0.924 & 0.887 & 0.761 & \textbf{1.000} \\
& \textbf{CAMT5$_\text{base}$}  & \textbf{0.369} & \textbf{0.951} & \textbf{0.931} & \textbf{0.843} & \textbf{1.000} \\
\bottomrule
\end{tabular}
\vspace{-0.1in}
\caption{
Performance on ChEBI-20 \citep{edwards2021text2mol} grouped by number of atoms in molecules. $\dagger$ denotes that the model is trained with the same training configuration, e.g., training dataset, as ours.
}
\label{tab:atomcount_results}
\end{table}

%% file: tables/atom_level_description.tex
\begin{table}[ht]
\centering
\begin{tabular}{l|l|ccccc}
\toprule
\textbf{Description} & \textbf{Model} & \textbf{Exact} $\uparrow$ & \textbf{MACCS} $\uparrow$ & \textbf{RDK} $\uparrow$ & \textbf{Morgan} $\uparrow$ & \textbf{Validity} $\uparrow$ \\
\midrule
\multirow{3}{*}{`chlor'}
& MolT5$_\text{base}^\dagger$   & 0.244 & 0.781 & 0.677 & 0.589 & 0.944 \\
& BioT5$_\text{base}^\dagger$   & 0.216 & 0.738 & 0.638 & 0.524 & \textbf{1.000} \\
& \textbf{CAMT5$_\text{base}$} & \textbf{0.258} & \textbf{0.793} & \textbf{0.702} & \textbf{0.603} & \textbf{1.000} \\
\midrule
\multirow{3}{*}{`fluoro'}
& MolT5$_\text{base}^\dagger$   & 0.223 & 0.790 & 0.709 & 0.610 & 0.961 \\
& BioT5$_\text{base}^\dagger$   & 0.204 & 0.738 & 0.644 & 0.523 & \textbf{1.000} \\
& \textbf{CAMT5$_\text{base}$} & \textbf{0.262} & \textbf{0.816} & \textbf{0.715} & \textbf{0.622} & \textbf{1.000} \\
\midrule
\multirow{3}{*}{`phospho'}
& MolT5$_\text{base}^\dagger$   & 0.371 & 0.911 & 0.872 & 0.824 & 0.976 \\
& BioT5$_\text{base}^\dagger$   & 0.510 & 0.910 & 0.854 & 0.799 & \textbf{1.000} \\
& \textbf{CAMT5$_\text{base}$} & \textbf{0.614} & \textbf{0.952} & \textbf{0.916} & \textbf{0.865} & \textbf{1.000} \\
\midrule
\multirow{3}{*}{`sulf'}
& MolT5$_\text{base}^\dagger$   & 0.387 & 0.856 & 0.779 & 0.707 & 0.955 \\
& BioT5$_\text{base}^\dagger$   & 0.300 & 0.831 & 0.744 & 0.632 & \textbf{1.000} \\
& \textbf{CAMT5$_\text{base}$} & \textbf{0.424} & \textbf{0.882} & \textbf{0.825} & \textbf{0.740} & \textbf{1.000} \\
\bottomrule
\end{tabular}
\vspace{-0.1in}
\caption{
Performance on ChEBI-20 \citep{edwards2021text2mol} containing specific atom-level descriptions. $\dagger$ denotes that the model is trained with the same training configuration, e.g., training dataset, as ours.
}
\label{tab:atom_level_grouped}
\end{table}

%% file: tables/frag_ablation.tex
\begin{table}[ht]
\centering
\begin{tabular}{l|ccccc}
\toprule
\textbf{Tokenization} & \textbf{Exact} $\uparrow$ & \textbf{MACCS} $\uparrow$ & \textbf{RDK} $\uparrow$ & \textbf{Morgan} $\uparrow$ & \textbf{Validity} $\uparrow$ \\
\midrule
t-SMILES  \citep{wu2024t}  & 0.025 & 0.700 & 0.636 & 0.475 & 0.997 \\
BRICS \citep{degen2008art} & 0.216 & 0.808 & 0.765 & 0.633 & 1.000 \\ \midrule
\textbf{Ours} & \textbf{0.391} & \textbf{0.874} & \textbf{0.827} & \textbf{0.727} & \textbf{1.000} \\
\bottomrule
\end{tabular}
\vspace{-0.1in}
\caption{
Comparison of tokenization strategies on ChEBI-20 \citep{edwards2021text2mol} using the models derived from T5-small \citep{raffel2020exploring}.
}
\label{tab:frag_ablation}
\end{table}

%% file: tables/search_ablation.tex
\begin{table}[ht]
\centering
\begin{tabular}{l|ccccc}
\toprule
\textbf{Algorithm}& \textbf{Exact} $\uparrow$ & \textbf{MACCS} $\uparrow$ & \textbf{RDK} $\uparrow$ & \textbf{Morgan} $\uparrow$ & \textbf{Validity} $\uparrow$ \\
\midrule
Breadth-First Search (BFS) & 0.368 & 0.858 & 0.808 & 0.707 & 1.000 \\ \midrule
\textbf{Depth-First Search (DFS)} & \textbf{0.391} & \textbf{0.874} & \textbf{0.827} & \textbf{0.727} & \textbf{1.000} \\
\bottomrule
\end{tabular}
\vspace{-0.1in}
\caption{
Ablation of linearization algorithms in our tokenization strategy on ChEBI-20 \citep{edwards2021text2mol} using the models derived from T5-small \citep{raffel2020exploring}.
}
\label{tab:search_ablation}
\end{table}

%% file: tables/importance_objective_ablation.tex
\begin{table}[ht]
\centering
\begin{tabular}{l|ccccc}
\toprule
\textbf{Importance} & \textbf{Exact} $\uparrow$ & \textbf{MACCS} $\uparrow$ & \textbf{RDK} $\uparrow$ & \textbf{Morgan} $\uparrow$ & \textbf{Validity} $\uparrow$ \\
\midrule
Frequency of atoms     & 0.390 & 0.867 & 0.816 & 0.719 & \textbf{1.000} \\
Frequency of motifs  & 0.281 & 0.811 & 0.758 & 0.638 & \textbf{1.000} \\ \midrule
\textbf{Number of atoms} & \textbf{0.391} & \textbf{0.874} & \textbf{0.827} & \textbf{0.727} & \textbf{1.000} \\
\bottomrule
\end{tabular}
\vspace{-0.1in}
\caption{
Ablation on the training objective in importance-based training on ChEBI-20 \citep{edwards2021text2mol} using the models derived from T5-small \citep{raffel2020exploring}.
}
\label{tab:importance_ablation}
\end{table}

%% file: tables/ensemble_ablation_acc.tex
\begin{table}[ht]
\centering
\begin{tabular}{ccc|ccccc}
\toprule
\textbf{MolT5} & \textbf{BioT5} & \textbf{CAMT5 (Ours)} & \textbf{Exact} $\uparrow$ & \textbf{MACCS} $\uparrow$ & \textbf{RDK} $\uparrow$ & \textbf{Morgan} $\uparrow$ & \textbf{Validity} $\uparrow$ \\
\midrule
 \textcolor{darkblue}\cmark&\textcolor{darkblue}\cmark&  \textcolor{SJRed}\xmark& 0.443 & 0.889 & 0.841 & 0.760 & \textbf{1.000} \\
 \textcolor{darkblue}\cmark& \textcolor{SJRed}\xmark & \textcolor{darkblue}\cmark&  0.455 & 0.891 & 0.849 & 0.768 & \textbf{1.000} \\
  \textcolor{SJRed}\xmark& \textcolor{darkblue}\cmark&\textcolor{darkblue}\cmark& {0.462} & \textbf{0.902} & {0.857} & {0.772} & \textbf{1.000} \\ \midrule
 \textcolor{darkblue}\cmark& \textcolor{darkblue}\cmark& \textcolor{darkblue}\cmark& \textbf{{0.472}} & \textbf{{0.902}} & \textbf{{0.860}} & \textbf{{0.781}} & \textbf{{1.000}}\\

\bottomrule
\end{tabular}
\vspace{-0.1in}
\caption{
Performance of ensemble with different combinations of models on ChEBI-20 \citep{edwards2021text2mol}. We report the results based on the best-performing model of $\text{MolT5, BioT5, \Algname}$ in Table~\ref{tab:chebi_main}, respectively.
}
\label{tab:ensemble_accuracy}
\end{table}

%% file: tables/data_efficient_generation.tex
\begin{table}[ht]
\centering
\begin{tabular}{l|cccccc}
\toprule
\textbf{Importance} & \textbf{Active.} $\uparrow$ & \textbf{FCD} $\uparrow$ & \textbf{NSPDK} $\uparrow$ & \textbf{Valid.} $\uparrow$ & \textbf{Unique.} $\uparrow$ & \textbf{Novelty} $\uparrow$\\
\midrule
MolT5 \citep{edwards2022translation}  & 11.2 & 18.7 & 0.020 & 70.4 & 87.2 & \textbf{100} \\
BioT5 \citep{pei2023biot5}  & 11.6 & 17.0 & 0.019 & \textbf{100} & \textbf{96.8} & \textbf{100} \\ \midrule
\textbf{\Algname (Ours)} & \textbf{12.0} & \textbf{16.3} & \textbf{0.018} & \textbf{100} & {96.4} & \textbf{100} \\
\bottomrule
\end{tabular}
\vspace{-0.1in}
\caption{
Results on data-efficient generation \citep{kim2024data} on the HIV dataset \citep{wu2018moleculenet}.
}
\label{tab:data_effficient_generation}
\end{table}

%% file: tables/statistical_analysis.tex
\begin{table}[ht]
\centering
\begin{tabular}{l|ccccc}
\toprule
\textbf{Model} & \textbf{Exact} $\uparrow$ & \textbf{MACCS} $\uparrow$ & \textbf{RDK} $\uparrow$ & \textbf{Morgan} $\uparrow$ & \textbf{Validity} $\uparrow$ \\
\midrule
MolT5 \citep{edwards2021text2mol}     & 0.278\stdv0.009 & 0.836\stdv0.006 & 0.779\stdv0.005 & 0.696\stdv0.005 & 0.946\stdv0.004 \\
BioT5 \citep{pei2023biot5}  & 0.324\stdv0.010 & 0.841\stdv0.002 & 0.771\stdv0.003 & 0.658\stdv0.004 & \textbf{1.0}\stdv0.0 \\ \midrule
\textbf{\Algname (Ours)} & \textbf{0.388}\stdv0.003 & \textbf{0.871}\stdv0.003 & \textbf{0.823}\stdv0.004 & \textbf{0.723}\stdv0.003 & \textbf{1.0}\stdv0.0 \\ 
\bottomrule
\end{tabular}
\vspace{-0.1in}
\caption{
Comparison of the mean and standard deviation values based on the text-to-molecule models derived from T5-small \citep{raffel2020exploring}. The results are calculated over 3 runs with different seeds.
}
\label{tab:statistical_analysis}
\end{table}

%% file: tables/computational_cost.tex
\begin{table}[ht]
\centering
\begin{tabular}{l|ccc}
\toprule
\textbf{Cost}& \textbf{Fine-tuning cost (hrs)} & \textbf{Memory (GB)} & \textbf{Inference cost (sec)}  \\
\midrule
MolT5 \citep{edwards2021text2mol} & 20 & 3.95 & 0.65 \\ 
BioT5 \citep{pei2023biot5} & 19 & 3.95 & 0.61 \\ \midrule
\textbf{\Algname (Ours)} & \textbf{15} & \textbf{2.79} & \textbf{0.30} \\
\bottomrule
\end{tabular}
\vspace{-0.1in}
\caption{
Comparison of computational costs among the models derived from T5-base \citep{raffel2020exploring}.
}
\label{tab:computational_cost}
\end{table}